\documentclass{article}

\usepackage{microtype}
\usepackage{graphicx}
\usepackage{subfigure}
\usepackage{booktabs}
\usepackage{subcaption}
\usepackage{rotating}

\usepackage[hyphens]{url}
\usepackage{hyperref}
\hypersetup{colorlinks=true,breaklinks=true}

\usepackage[accepted]{icml2024}

\usepackage{amsmath}
\usepackage{amssymb}
\usepackage{mathtools}
\usepackage{amsthm}
\usepackage{bm}

\usepackage[capitalize,noabbrev]{cleveref}

\theoremstyle{plain}

\theoremstyle{definition}

\theoremstyle{remark}

\usepackage[textsize=tiny]{todonotes}

\icmltitlerunning{When Representations Align: Universality in Representation Learning Dynamics}

\begin{document}

\twocolumn[

\icmltitle{When Representations Align:\\ Universality in Representation Learning Dynamics}

\icmlsetsymbol{equal}{*}

\begin{icmlauthorlist}
\icmlauthor{Loek van Rossem}{gatsby}
\icmlauthor{Andrew M. Saxe}{gatsby,swc}
\end{icmlauthorlist}

\icmlaffiliation{gatsby}{Gatsby Computational Neuroscience Unit, University College London}
\icmlaffiliation{swc}{Sainsbury Wellcome Centre, University College London}

\icmlcorrespondingauthor{Loek van Rossem}{loek.rossem.22@ucl.ac.uk}

\icmlkeywords{Universality, Representation Learning, Deep Learning Theory}

\vskip 0.3in
]

\printAffiliationsAndNotice{}
\begin{abstract}
Deep neural networks come in many sizes and architectures. The choice of architecture, in conjunction with the dataset and learning algorithm, is commonly understood to affect the learned neural representations. Yet, recent results have shown that different architectures learn representations with striking qualitative similarities. Here we derive an effective theory of representation learning under the assumption that the encoding map from input to hidden representation and the decoding map from representation to output are arbitrary smooth functions. This theory schematizes representation learning dynamics in the regime of complex, large architectures, where hidden representations are not strongly constrained by the parametrization. We show through experiments that the effective theory describes aspects of representation learning dynamics across a range of deep networks with different activation functions and architectures, and exhibits phenomena similar to the ``rich" and ``lazy" regime. While many network behaviors depend quantitatively on architecture, our findings point to certain behaviors that are widely conserved once models are sufficiently flexible. 
\end{abstract}

\section{Introduction}
One of the major challenges currently faced in the theory of deep learning is the issue of scalability. Although exact solutions to the learning dynamics have been found for some simple networks \cite{saad_exact_1995,saxe_exact_2014,mei_mean_2018,arora_exact_2019,saxe_neural_2022}, under these analyses, any small change to the network architecture requires one to significantly amend the analysis. A solution to the learning dynamics of one architecture does not necessarily transfer to another architecture, a rather inconvenient fact considering the large range of different models available. Moreover, state of the art models have a complexity far exceeding what can be reasonably solved analytically.

Studies have shown that although the exact representation between different neural networks may differ, some important aspects of the computational structure they learn are universal \cite{maheswaranathan_universality_2019,mcmahan_learning_2021,huang_semantic_2021}. Even for certain areas in the brain, similarities in representational structure have been found with deep neural networks trained on natural image data \cite{yamins_performance-optimized_2014,yamins_using_2016,saxe_if_2021,conwell_what_2023,bracci_representational_2023}. The data used in learning may play a larger role than the implementation details of the model, a fact which has been empirically demonstrated in vision transformers and convolutional neural networks \cite{bouchacourt_grounding_2021,conwell_what_2023} and studied theoretically \cite{goldt_modelling_2020}. Additionally, neural scaling laws describe a formal relationship between model or dataset size and performance that appears to hold without making direct reference to architecture \cite{kaplan_scaling_2020,hoffmann_training_2022}. More recently, it has been found that an alternative architecture performs roughly the same as the transformer, given a similar parameter count \cite{wang_pretraining_2023}. These observations together license universal theories: mathematical descriptions of learning that sidestep certain implementation details in order to model behavior common among architectures.

Although a universal theory by necessity cannot completely capture all details of any particular architecture, it has distinct advantages. Derived results may apply to many different learning systems. In particular, this could include highly complicated machine learning models as well as the brain, systems which are difficult to study theoretically otherwise. Furthermore, such an abstract approach forces us to ignore many details not essential to describing specific learning behaviors. This may aid in deriving results, as it reduces the mathematical complexity of the theory. It will also focus our attention to only the most crucial concepts needed to understand learning dynamics, providing a better sense of how we should be thinking about these highly complex systems.

Here we propose an approach to modelling universal representation learning, striving to provide conceptual explanations for phenomena universally observed in learning systems. We seek to summarize known results in deep learning within a single framework. Our main contributions are as follows:
    
\begin{itemize}
\item In \cref{sec:the_model}, we derive an effective theory for the interaction between two datapoints with nearby representations during training in the expressive regime, where the neural network is large and complex enough to not be strongly constrained by its parametrization.
\item In \cref{sec:universal_dynamics}, we demonstrate the existence of universal behavior within representational learning dynamics by showing that the derived theory describes the dynamics of a range of deep networks with varying activation functions and architectures.
\item In \cref{sec:structured_representations}, we look at the final representational structure predicted by the theory, and find two qualitatively different regimes, one based on random initializations, and one based on structure in the data. We run experiments on deep neural networks trained on toy datasets, and find similar representations as predicted by the theory.
\end{itemize}

\subsection{Related Work}

\paragraph{Exact solutions in simple architectures.}
Considerable progress has been made in the theoretical analysis of deep linear neural networks, for example, the loss landscape is well understood \cite{baldi_neural_1989,kawaguchi_deep_2016}, and exact solutions have been obtained for specific initial conditions \cite{saxe_exact_2014,lampinen_analytic_2019,braun_exact_2022,pesme_saddle--saddle_2023}. This is in contrast to other architectures, for which only few exact analyses of learning dynamics exists, see e.g. \cite{seung_statistical_1992,saad_exact_1995,mei_mean_2018,saxe_neural_2022}. Although some of these results derived for linear networks reflect behavior in more complex architectures, there is no precise mathematical connection in the dynamics of linear and nonlinear neural networks. In this work, we will attempt to make this relationship more explicit.

\paragraph{The neural tangent kernel.}
A notable exception in terms of universal solutions is that of the neural tangent kernel literature \cite{lee_wide_2019,jacot_neural_2020,golikov_neural_2022}, which does provide exact solutions applicable to a wide range of models. The limitation here is that the neural tangent kernel solutions are only valid for infinitely wide neural networks at a specific weight initialization scale \cite{woodworth_kernel_2020}. In particular, the initialization regime used is one of relatively high weights, where the dataset does not impact the representation \cite{chizat_lazy_2020,flesch_orthogonal_2022}. We seek a similar universal theory, applicable to the feature learning regime, where the dataset controls the representation.

\paragraph{Implicit biases in gradient descent.}
A large body of work has investigated gradient descent as a source of generalization performance in deep neural networks, as opposed to explicit regularization and inductive biases in the architecture \cite{neyshabur_search_2015,gunasekar_implicit_2018,chizat_implicit_2020,soudry_implicit_2022}. This work is mainly either empirical or focused on analyses for examples of specific architectures. Here instead we consider a modeling approach, aiming to provide intuition into the universal nature of implicit bias and observed similarities between representational structures across architectures.

\paragraph{Local elasticity.}
One motivating factor for the method used here is the empirical phenomenon of local elasticity \cite{he_local_2020,zhang_imitating_2021}, observed for non-linear neural networks performing classification. A model is said to exhibit local elasticity in the case that updating one feature vector does not significantly perturb feature vectors dissimilar to it. This allows us to consider neural network learning dynamics as a local process, suggesting that the local interaction between representations may already provide insight into the behavior of the entire dataset.

\paragraph{Physics inspired approaches.}
The approach to modelling a two point interacting between datapoints taken here is inspired by intuitions from physics surrounding interacting particles. Physics-inspired particle approaches are prominent in the deep learning theory \cite{seung_statistical_1992, rotskoff_neural_2018, rotskoff_trainability_2022, baek_geneft_2024}, most notably the application of mean field theory \cite{saul_mean_1996, poole_exponential_2016, xiao_dynamical_2018, mei_mean_2018, mei_mean-field_2019, pandey_mean_2019}.

\section{The Model}
\label{sec:the_model}
The approach we will be taking here is motivated by universal approximation theorems \cite{hornik_multilayer_1989,barron_universal_1993,csaji_approximation_2001}, which state that given enough parameters, non-linear neural networks can learn to approximate any arbitrary smooth function. We can thus consider the process of training a neural network as an optimization of a smooth function to fit the data. Constraints coming from the architecture of the neural network may affect the dynamics of the optimization process, but since here we are interested in modelling these function dynamics without reference to the details of the architecture, we will choose to ignore this.

\subsection{Function Dynamics}

More explicitly, let us consider some unspecified deep neural network $f_{\theta} : X \to Y$, that is trained on a dataset $\mathcal{D} = \{(x_i, y_i)\}_{i=1}^N, \forall x_i \in X, \forall y_i \in Y$. During training its set of parameters $\theta$ is optimized to reduce the mean squared error loss function:
\begin{equation}
    L(f_{\theta})=\frac{1}{2} \langle ||f_{\theta}(x_i)-y_i||^2\rangle_{\mathcal{D}}
    ,
    \label{eq:mean_squared_error}
\end{equation}
where $\langle \cdot \rangle_\mathcal{D}$ is the average over the dataset $\mathcal{D}$.

Since we are interested in studying the dynamics of the representational space, we will split $f_{\theta}$ into two smooth maps, an encoder map $h_{\phi}: X \to H$ which assigns to each input in $X$ a representation in some intermediate layer $H$ in the neural network, and a decoder map $y_{\psi}: H \to Y$ which assigns a predicted output to each possible representation in $H$. We may then write the loss in \cref{eq:mean_squared_error} as:
\begin{equation}
    L(h_{\phi}, y_{\psi})=\frac{1}{2} \langle ||y_{\psi}(h_{\phi}(x_i))-y_i||^2\rangle_{\mathcal{D}}
    .
\end{equation}
The dynamics on these maps follow from applying the gradient descent update rule to the parameters $\phi$ and $\psi$ defining them:
\begin{equation}
    \Delta \phi = -\eta \nabla_{\phi} L \quad \text{and} \quad \Delta \psi = -\eta \nabla_{\psi} L
    ,
\end{equation}
where $\eta$ is the learning rate. For sufficiently small $\eta$ we can take the continuous time limit
\begin{equation}
    \frac{\mathrm{d}}{\mathrm{d}t} \phi = -\frac{1}{\tau} \nabla_{\phi} L \quad \text{and} \quad  \frac{\mathrm{d}}{\mathrm{d}t} \psi = -\frac{1}{\tau} \nabla_{\psi} L
    ,
    \label{eq:gradient_descent}
\end{equation}
where $\tau := \frac{1}{\eta}$.
The problem with \cref{eq:gradient_descent} is that it explicitly depends on the parametrization of the neural network, whereas a universal approach requires ignoring such implementation details. Ideally we would like to optimize the loss $L$ directly with respect to the maps $h_{\phi}$ and $y_{\psi}$, since if the model has high \textit{expressivity}, i.e. the architecture is large and complex enough, perhaps it has enough freedom to behave as an arbitrary smooth map. However, it is unclear how this can be achieved mathematically\footnote{One may be tempted to try replacing the right hand side of \cref{eq:gradient_descent} with functional derivatives of $L$ with respect to $h_{\phi}$ and $y_{\psi}$, however this will result in a term involving the gradient of $y_{\psi}$, and the dynamics of this gradient explicitly depend on $\psi$.}.

\subsection{Two Point Interaction}

In general, the modelling issue of parametrization dependent dynamics does not appear to have a resolution. So let us instead restrict ourselves to a simpler case, where the only property we do know about the unspecified network, the smoothness of the maps $h_{\phi}$ and $y_{\psi}$, can be exploited. During training the representations of different datapoints move around in the hidden space $H$, and may get near each other and interact. Understanding such an interaction on its own may already provide some insight into the representational learning behavior of the model on the full dataset.

To model this two point interaction, let us consider a dataset $\mathcal{D} = \{(x_1, y_1), (x_2, y_2)\}$. If the representations $h_{\phi}(x_1)$ and $h_{\phi}(x_2)$ are close enough, then by smoothness it is reasonable to a take a linear approximation of $h_{\phi}$ and $y_{\psi}$ around the mean of the two points:
\begin{equation}
    \begin{split}
        h_{\phi}(x_{i})&\approx h_{\phi}(\frac{x_{2}+x_{1}}{2})+\frac{1}{2}D_{h}(x_{i}-x_{\neg i})\\
y_{\psi}(x_{i})&\approx y_{\psi}(h_{\phi}(\frac{x_{2}+x_{1}}{2}))+\frac{1}{2}D_{y}(D_{h}(x_{i}-x_{\neg i}))
    \end{split}
    ,
    \label{eq:linear_approximation}
\end{equation}
where $D_{h}$ and $D_{y}$ are the Jacobian matrices of $h_{\phi}$ and $y_{\psi}$ respectively.

Presumably, as gradient descent optimizes $\phi$ and $\psi$ in order to reduce the loss, at the two point interaction this will effectively optimize the linearization parameters $D_{h}$, $D_{y}$ and $\bar{y} := y_{\psi}(h_{\phi}(\frac{x_{2}+x_{1}}{2}))$ in order to minimize the loss\footnote{Potentially $h_{\phi}(\frac{x_{2}+x_{1}}{2})$ should also be considered as a linearization parameter, but it will turn out to not play a role in the following analysis.}. Assuming expressivity, i.e. ``enough freedom" in the architecture such that there are no additional constraints on the dynamics, we can model the dynamics by applying gradient descent directly to the linearization parameters, i.e.
\begin{equation}
    \begin{split}
        \frac{\mathrm{d}}{\mathrm{d}t} \bar{y} &= -\frac{1}{\tau_{\bar{y}}}\frac{\partial L}{\partial \bar{y}}\\
        \frac{\mathrm{d}}{\mathrm{d}t}  D_h &= -\frac{1}{\tau_h}\frac{\partial L}{\partial D_h}\\
        \frac{\mathrm{d}}{\mathrm{d}t}  D_y &= -\frac{1}{\tau_y}\frac{\partial L}{\partial D_y}
        ,
    \end{split}
    \label{eq:universal_dynamics}
\end{equation}
where $1/\tau_{\bar{y}}, 1/\tau_{h}$ and $1/\tau_{y}$ are effective learning rates.

Note that these dynamics are not necessarily equal to the actual dynamics on the linearization parameters induced by the model parameters $\phi$ and $\psi$. \cref{eq:universal_dynamics} is the main modelling assumption, and intended as an \textit{effective theory} in the regime of large complex architectures, where behavior is unconstrained by their parametrization. The details of how $h_{\phi}$ and $y_{\psi}$ are parameterized determine the effective learning rates $1/\tau_{\bar{y}}, 1/\tau_{h}$ and $1/\tau_{y}$, but are otherwise not present in the effective theory.

In fact, if one were to interpret the linearization parameters as model weights, then the effective theory dynamics are mathematically the same as that of a two layer linear network with bias and differing learning rates for each of the layers. However, we are not using a two layer linear network to model the neural network here, instead we are locally modeling a two point interaction.

\subsection{Reduction to a 3-dimensional System}

To simplify the dynamics, we study solutions where the representations of the two datapoints either move together or apart but do not rotate. Strikingly, as derived in \cref{sec:reduction_to_a_3-dimensional_system}, this results in a self-contained system of three scalar variables:
\begin{equation}
    \begin{split}
        \frac{\mathrm{d}}{\mathrm{d}t} ||dh||^2 =& -\frac{1}{\tau_h} ||x_2-x_1||^2 w\\
        \frac{\mathrm{d}}{\mathrm{d}t} ||dy||^2 =& -w(\frac{1}{\tau_y} ||dh||^2 +\frac{1}{\tau_h} ||x_2-x_1||^2 \frac{||dy||^2}{||dh||^2}) \\
        \frac{\mathrm{d}}{\mathrm{d}t} w =& -\frac{1}{\tau_y} \frac{1}{2}(3w-||dy||^2+||y_2-y_1||^2)||dh||^2\\
        &-\frac{1}{\tau_h}\frac{1}{2} ||x_2-x_1||^2 \frac{(||dy||^2+w)w}{||dh||^2}
        ,
    \end{split}
    \label{eq:dynamics}
\end{equation}
where
\begin{equation}
    \begin{split}
        ||dh||^2:=&||D_h(x_2-x_1)||^2\\
        ||dy||^2:=&||D_y D_h(x_2-x_1)||^2\\
        w:=&||dy||^2-(D_y D_h(x_2-x_1))^{\top}(y_2-y_1)
        .
    \end{split}
\end{equation}

\begin{figure}[ht]
\vskip 0.2in
\begin{center}
\centerline{\includegraphics[width=\columnwidth]{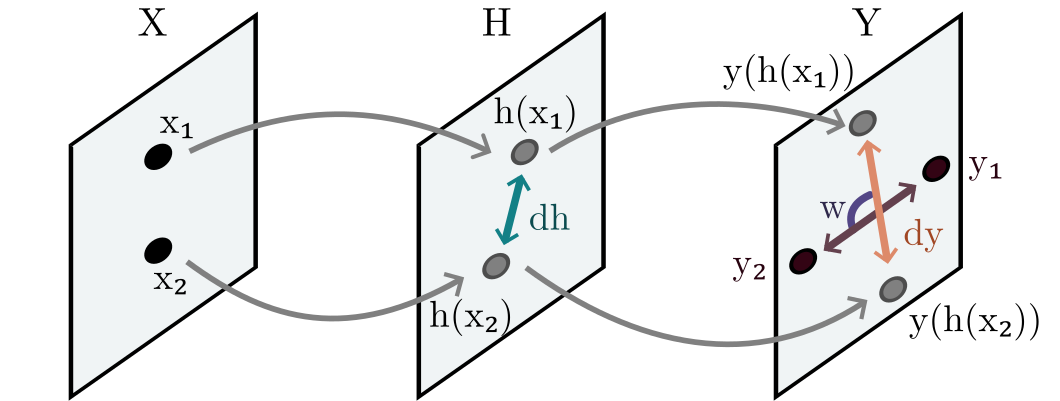}}
\caption{Overview of the effective theory for the two point interaction. For two datapoints $x_1$ and $x_2$ self-contained dynamics are defined on their representational difference $||dh||$, predicted output difference $||dy||$, and output alignment $w$.}
\label{fig:overview}
\end{center}
\vskip -0.2in
\end{figure}

We will find that this system still exhibits interesting behavior worth studying. \cref{fig:overview} illustrates the relevant variables of this system. The main quantity we care about here is the representational distance $||dh||$, as it may provide insight into what kind of representational structures are learned. The difference in predicted outputs $||dy||$ is useful for modelling loss curves. Finally, there is the additional variable $w$, which controls the representational velocity. This can be interpreted as an alignment as it only contains additional information about the angle between the predicted outputs and the target outputs. 

\begin{figure*}
  \includegraphics[width=\textwidth]{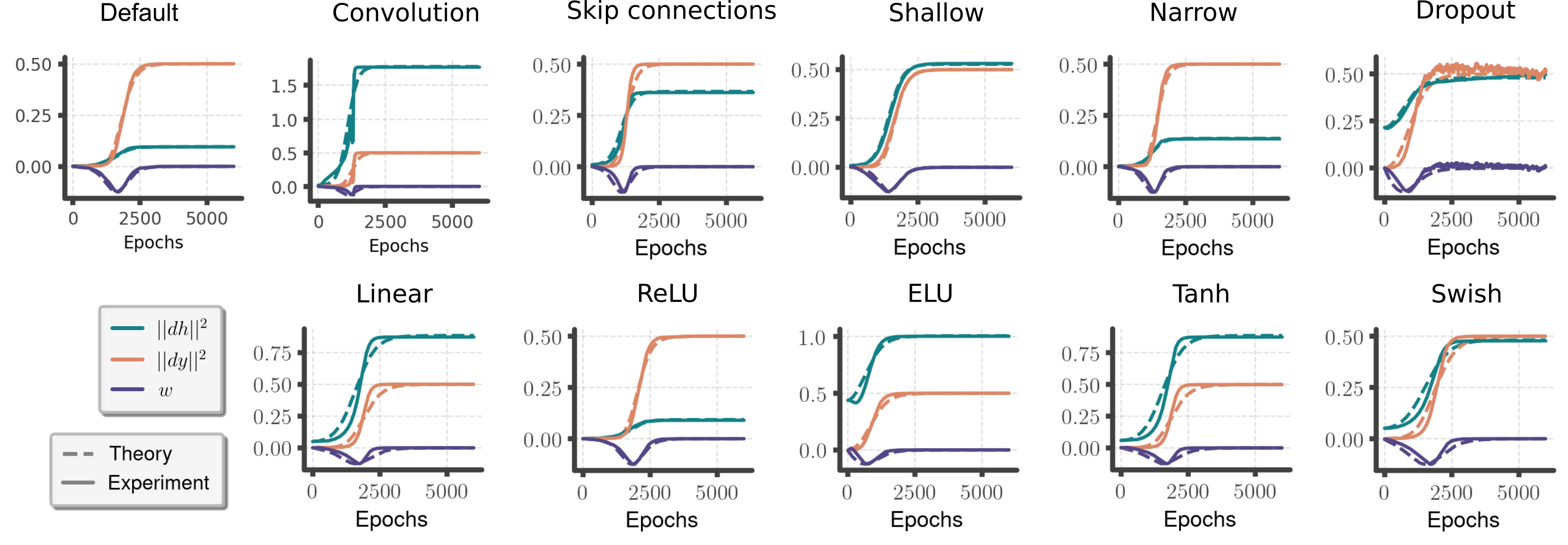}
  \caption{Universal learning dynamics among different architectures. The representational distance $||dh||^2$, prediction difference $||dy||^2$ and output alignment $w$ during training on a two point dataset amongst architectures with varying connectivity (top) and nonlinearities (bottom), matches the theory after fitting two constants. The architectures used are all variations of the default architecture and initialized at small weights so as to be in the expressive feature learning regime. Details for all experiments can be found in \cref{sec:experiment_details}.}
  \label{fig:traj_architectures}
\end{figure*}

\section{Universal Dynamics}
\label{sec:universal_dynamics}
The dynamics in \cref{eq:dynamics} contain no direct reference to the implementation details of the neural network, only two effective learning rates $1/\tau_h$ and $1/\tau_y$, which are unknown quantities and may differ across neural networks. Implementation details have essentially been abstracted away into these two constants. Given the right effective learning rates, solutions to these dynamics should model the learning dynamics of neural networks regardless of the architecture.

\subsection{Comparison of the Dynamics}
In \cref{fig:traj_architectures} we compare the learning dynamics of a neural network (20 layer, leaky ReLU, 500 units per layer) trained on a two point dataset to numerical solutions of the effective theory, as well as variations of this network with different architectures and nonlinearities. Despite only needing to fit two constants, the theory matches all three curves quite well for all networks. Note that this is not solely a result of having two degrees of freedom (see \cref{sec:alternative_dynamics}).

\subsection{Varying Initial Weights}
\label{sec:varying_initial_weights}
For small initial weights, the variables $||dh||, ||dy||$ and $w$ start small\footnote{There is an assumption here about the architecture. In principle, small weights could mean anything if nothing is know about how the smooth maps are parameterized, however, for typical neural network architectures these variables are small. }, so the right hand side of \cref{eq:dynamics} must be small too. Therefore, the dynamics are initially stagnant resulting in plateau-like behavior, whereas for large initial weights we have immediate exponential decay, as derived analytically in \cref{sec:lazy_learning_dynamics}. Such a transition from an initial plateau to immediate decay is observed experimentally, as can be seen in \cref{fig:traj_intialization} (top). Going to even larger initial weights, the representations start off far apart, so the linear approximation breaks down and the effective theory no longer matches the network.

\subsection{The Training Loss}

\begin{figure}[H]
\vskip 0.2in
\begin{center}
\centerline{\includegraphics[width=\columnwidth]{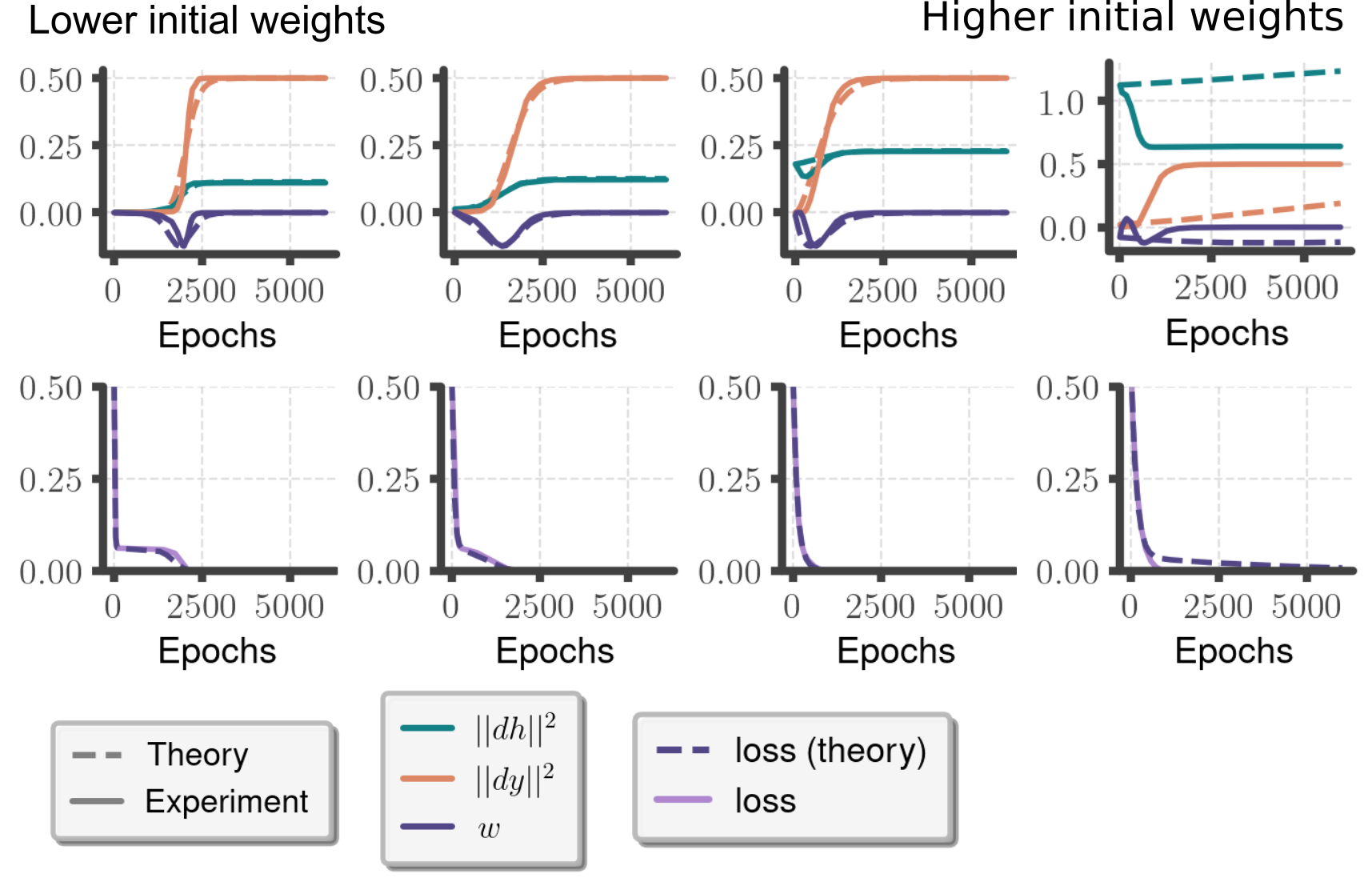}}
\caption{Dynamics of the 3-dimensional system (\textbf{top}) and training loss (\textbf{bottom}) at varying initial weights. The default architecture (20 fully connected layers, 500 units per layer, leaky ReLU) is trained on two datapoints and compared to the effective theory, after fitting two effective learning rates for the 3-dimensional system and one additional effective learning rate for the loss. Plateau-like behavior in the representational distance and loss can be seen at small initializations, but disappears at larger initial weights. At very high initial weights, when the representational distance starts off already large, the approximation breaks down as expected.}
\label{fig:traj_intialization}
\end{center}
\vskip -0.2in
\end{figure}

In \cref{sec:expression_for_the_loss_curve}, we derive an expression for the training loss as a function of $||dy||^2$ and $w$:
\begin{equation}
\begin{split}
    L=&\frac{1}{2}||(y(0)-\frac{y_2+y_1}{2}) e^{-\frac{t}{\tau_{\bar{y}}}}||^2\\
    & + \frac{1}{4}(w + \frac{1}{2} (||y_2-y_1||^2-||dy||^2)))
    ,
    \label{eq:loss_theory}
\end{split}
\end{equation}
where the third effective time constant $1/\tau_{\bar{y}}$ shows up. The first term in the expression is an exponential decay coming from fitting the output mean. The second term depends on $||dy||^2$ and $w$, which causes a plateau in the loss at small weights, due to the jump discussed in \cref{sec:varying_initial_weights}. The appearance of such a plateau at small weights is observed for the true dynamics, as can be seen in \cref{fig:traj_intialization} (bottom).

\subsection{Larger Datasets}
\begin{figure}[H]
\vskip 0.2in
\begin{center}
\centerline{\includegraphics[width=0.75\columnwidth]{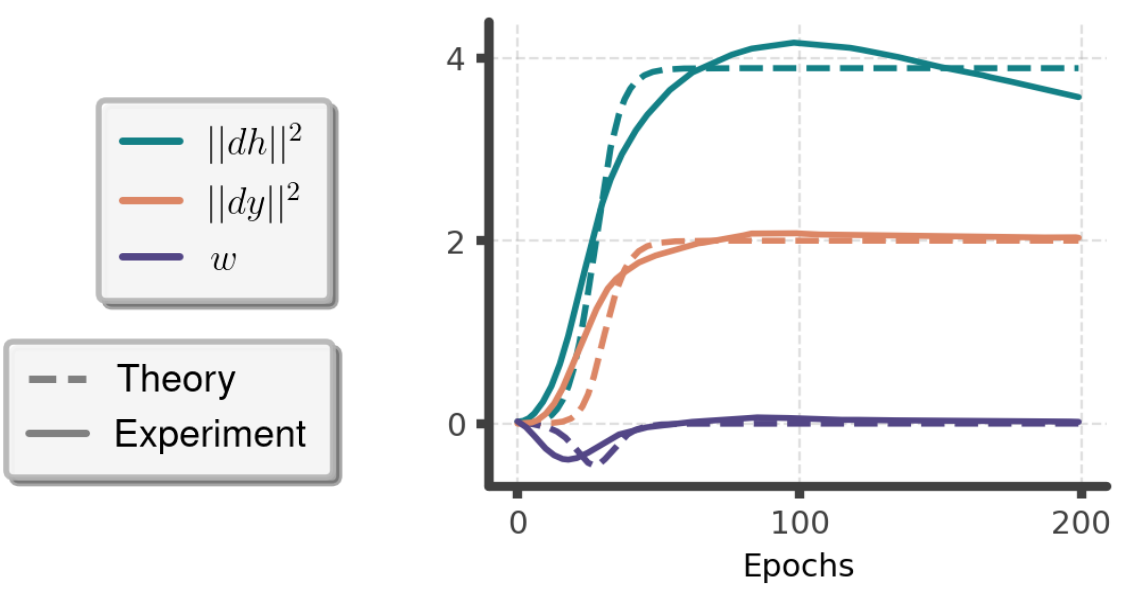}}
\caption{Learning dynamics of a randomly selected zero digit and one digit in MNIST compared against the theory after fitting two constants. The deep neural network has 4 fully connected layers, 100 units per layer, leaky ReLU activation and is initialized at small weights.}
\label{fig:traj_MNIST}
\end{center}
\vskip -0.2in
\end{figure}

We have only considered the interaction of two points in \textit{isolation}. Real datasets consist of many points, which may influence the behavior two interacting datapoints exhibit. To investigate the validity of the theory in this setting, we trained a model on the MNIST dataset, and tracked two distinguishable datapoints. The resulting dynamics (\cref{fig:traj_MNIST}) still qualitatively resemble the theory. 

\section{Structured Representations}
\label{sec:structured_representations}
If it is true that constraints coming from smoothness determine part of the behavior in neural network learning dynamics, what effect do these constraints have on the learned representation? To investigate this, we compute the final representational distance in the theory and compare it to neural networks trained on a few toy datasets.

\subsection{Lazy and Rich Learning Regimes}

\begin{figure*}[h]
  \includegraphics[width=\textwidth]{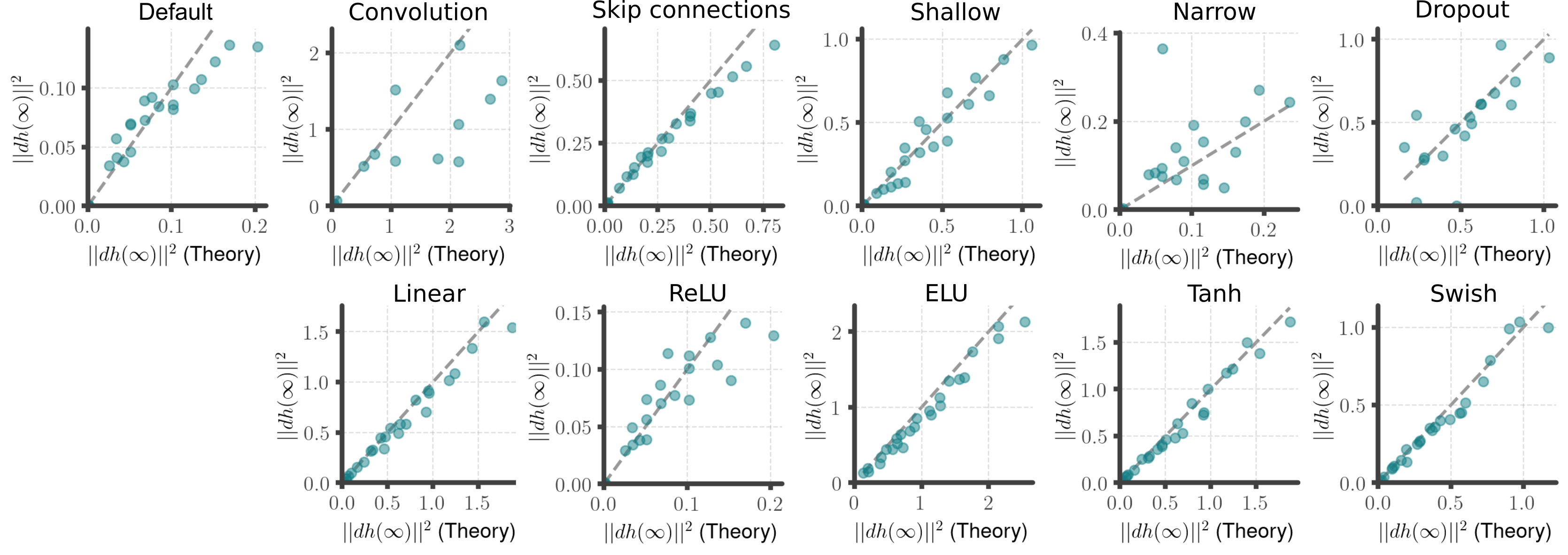}
  \caption{Structured learning at small initial weights. The final representational distance of two point experiments with varying input and target output distances compared against the theory (\cref{eq:final_distance}). Each dot represents a single trial, i.e. one training run. The effective learning rates for each architecture have been fit by averaging over 50 trials at \textit{fixed} input and target output distances.}
  \label{fig:structure_y}
\end{figure*}

We derive the fixed points of \cref{eq:dynamics} in \cref{sec:final_representational_distance}, using a reduction to a 2-dimensional system with dependence on initial conditions. We find that there is only one stable fixed point:
\begin{equation}
||dh||^2 = \frac{1}{2}A_{\text{high}} + \sqrt{\frac{1}{4}A^2_{\text{high}} + A^2_{\text{low}}}
,
\label{eq:final_distance}
\end{equation}
where
\begin{equation}
    \begin{split}
        A_{\text{high}} &= \left(\frac{||dh(0)||^2 }{||x_2-x_1||^2} - \frac{\tau_y}{\tau_h} \frac{||dy(0)||^2}{||dh(0)||^2}\right)||x_2-x_1||^2\\
        A_{\text{low}} & =\sqrt{\frac{\tau_y}{\tau_h}} ||x_2-x_1||\cdot ||y_2-y_1||
        ,
    \end{split}
\end{equation}
hence we expect this to be the final representational distance.

When the initial weights are large, i.e.
\begin{equation}
    \frac{||dh(0)||^2}{||x_2-x_1||^2}, \frac{||dy(0)||^2}{||dh(0)||^2} \gg 1
    ,
\end{equation}
the final representational distance converges to $A_\text{high}$, which depends on the \textit{input} structure of the data and the \textit{random initialization}. In the case that they are small
\begin{equation}
    \frac{||dh(0)||^2}{||x_2-x_1||^2}, \frac{||dy(0)||^2}{||dh(0)||^2} \ll 1
    ,
\end{equation}
it converges to $A_\text{low}$, which depends on the \textit{input} and \textit{output} structure of the data.

This separation between a random regime and a structured regime closely resembles the phenomenon of the ``rich" and ``lazy" learning that deep neural networks exhibit \cite{chizat_lazy_2020,flesch_rich_2021,flesch_orthogonal_2022,atanasov_onset_2022}. In particular, a crucial factor determining the regime is the initial weight scale \cite{woodworth_kernel_2020}.

\subsection{Rich Structure}
The representation at small initial weights is of particular interest, as it appears to be based on a structure found in the dataset, rather than formed by the random weight initialization. Note that again the only unknown quantities here are the two effective learning rates. By fitting these to the learning trajectories at a fixed choice of input and target output distances of the datapoints, we can extrapolate to predictions about the final representational distance for other distances, as shown in \cref{fig:structure_y}.

\begin{figure*}[h]
  \includegraphics[width=\textwidth]{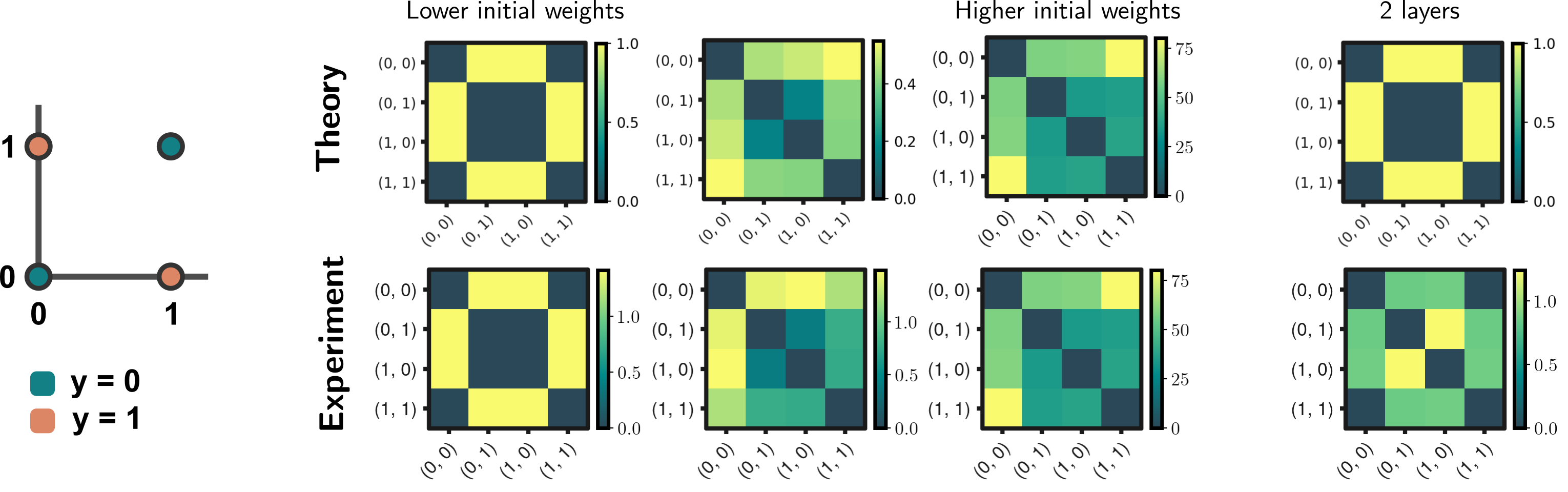}
  \caption{Representational structure in the XOR task. The XOR problem (\textbf{left}) consists of 4 datapoints with two input features and one alternating output feature. The learned hidden representational pair-wise distances match the theory prediction at varying initial weight scales for the default 20 layer network (\textbf{middle}), but not for a 2 layer network at small weights  (\textbf{right}).}
  \label{fig:XOR}
\end{figure*}

\subsection{Intuition}
\label{sec:intuition}
There is an intuition as to why smoothness constraints can result in these two different representational regimes. At large weights, representations are initially far apart, so the flexibility of the network allows the decoder to individually learn the correct output freely for each datapoint, without need of adjusting the representational structure. The dominating structure in the final representation is determined by the structure already present at initialization.

At small weights, representations are nearby and thus, due to smoothness constraints, they can no longer freely learn outputs without influencing each other. To illustrate this, consider the case where we have two datapoints with the same target output but only one already has the correct output prediction. There are two ways to learn the correct output for the other datapoint: (1) the decoder map can adjust at its representation and (2) the encoder map can move the representation towards the representation of the other datapoint, which due to continuity of the decoder map will align its output to that of the other datapoint, which already has the correct output. The second option therefore causes an effect on representations, where they move closer when their outputs agree. The opposite effect also holds: when two datapoints have different target outputs but the same representation, by continuity they cannot both have the correct output prediction. Their representations must move apart to fit the data.

Therefore, we see that at small weights there is a structured effect on the representation, dependent on the output data. This may help explain why overparameterized deep neural networks sometimes learn a representational structure, even without regularization. With a large dataset the interactions between datapoints may be numerous enough to dominate the representational learning behavior. In this case, \textit{the fastest way to learn the correct output is to learn a structured representation}, and gradient descent looks for the fastest way to minimize the loss. At small weights, the fastest way to learn may not be overfitting but instead structured learning, where representations due to their closeness benefit from each other.

\subsection{XOR Task}
To investigate the types of structures the theory is able to predict, we first consider the example of the XOR task. Here the aim is to learn a simple binary non-linear computation; there are only four datapoints, with two possible outputs, as illustrated in \cref{fig:XOR} (left). The theory predicts that at small initial weights, representations of datapoints with the same target output will merge, as $||y_2-y_1||$ is zero for such pairs. At large initial weights no such merging should occur. As can be seen in \cref{fig:XOR} (middle), this matches the learned representation of the large 20-layer default network. However, the theory fails to anticipate the merging of only one pair when the network is reduced to a two layer architecture, stressing the importance of the high model expressivity assumption (\cref{fig:XOR} (right)).

\subsection{Feature Collapse}
\begin{figure}[ht]
\vskip 0.2in
\begin{center}
\centerline{\includegraphics[width=\columnwidth]{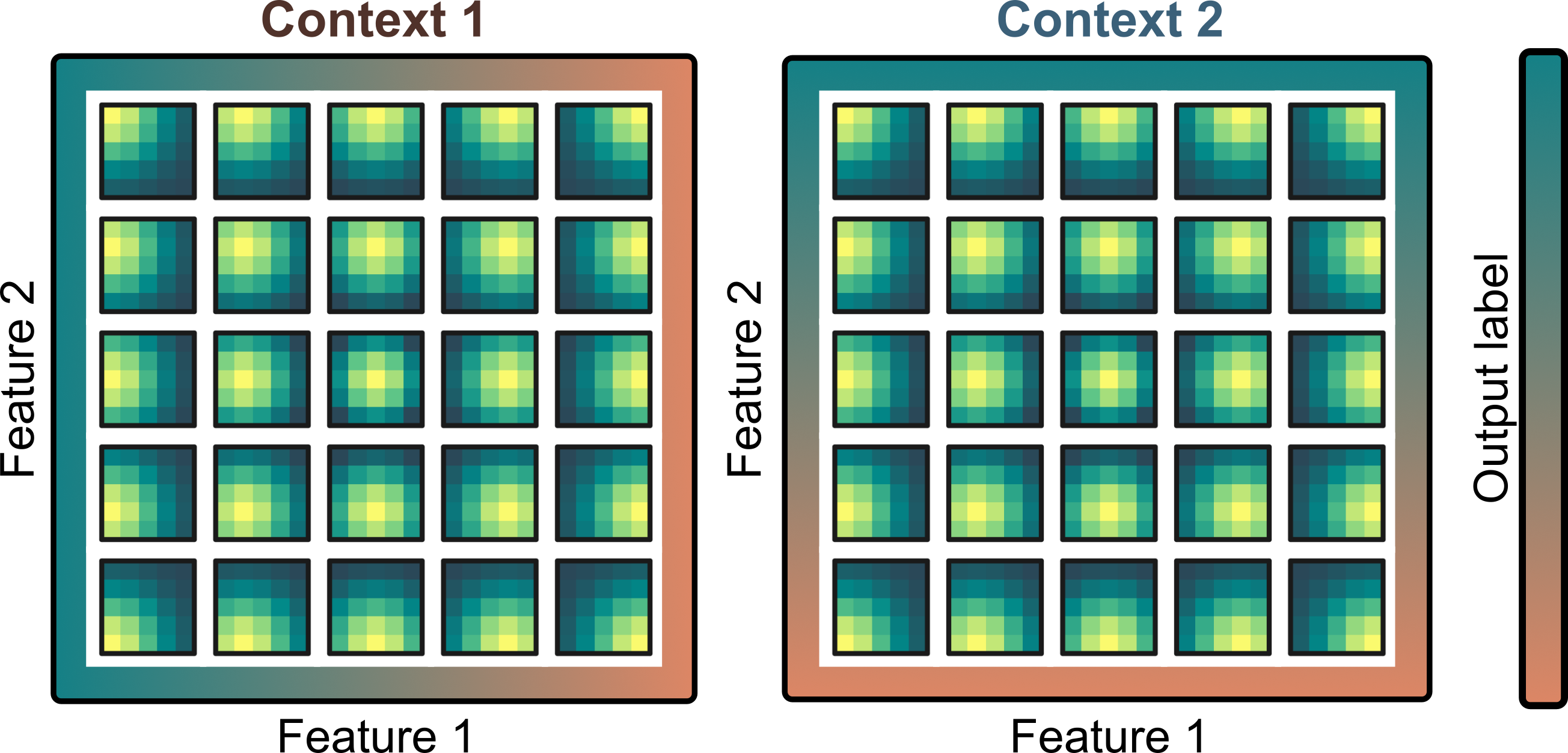}}
\caption{A toy dataset with two orthogonal features. The dataset consists of 2*30*30 datapoints, each 5*5 pixel images of Gaussian blobs, centered at varying x and y positions. A one hot encoded context variable is added, specifying whether the goal is to learn the feature 1 (x-position) or the feature 2 (y-position). }
\label{fig:feature_collapse_task}
\end{center}
\vskip -0.2in
\end{figure}

Next we consider an example of feature collapse explored in \cite{flesch_orthogonal_2022}. Here a network is trained on a dataset consisting of images of Gaussian blobs with varying x and y positions, as shown in \cref{fig:feature_collapse_task}. The task contains two contexts, one where the aim is to predict the x position of the blobs mean, and one where the aim is to predict its y position.

\begin{figure}[ht]
\vskip 0.2in
\begin{center}
\centerline{\includegraphics[width=\columnwidth]{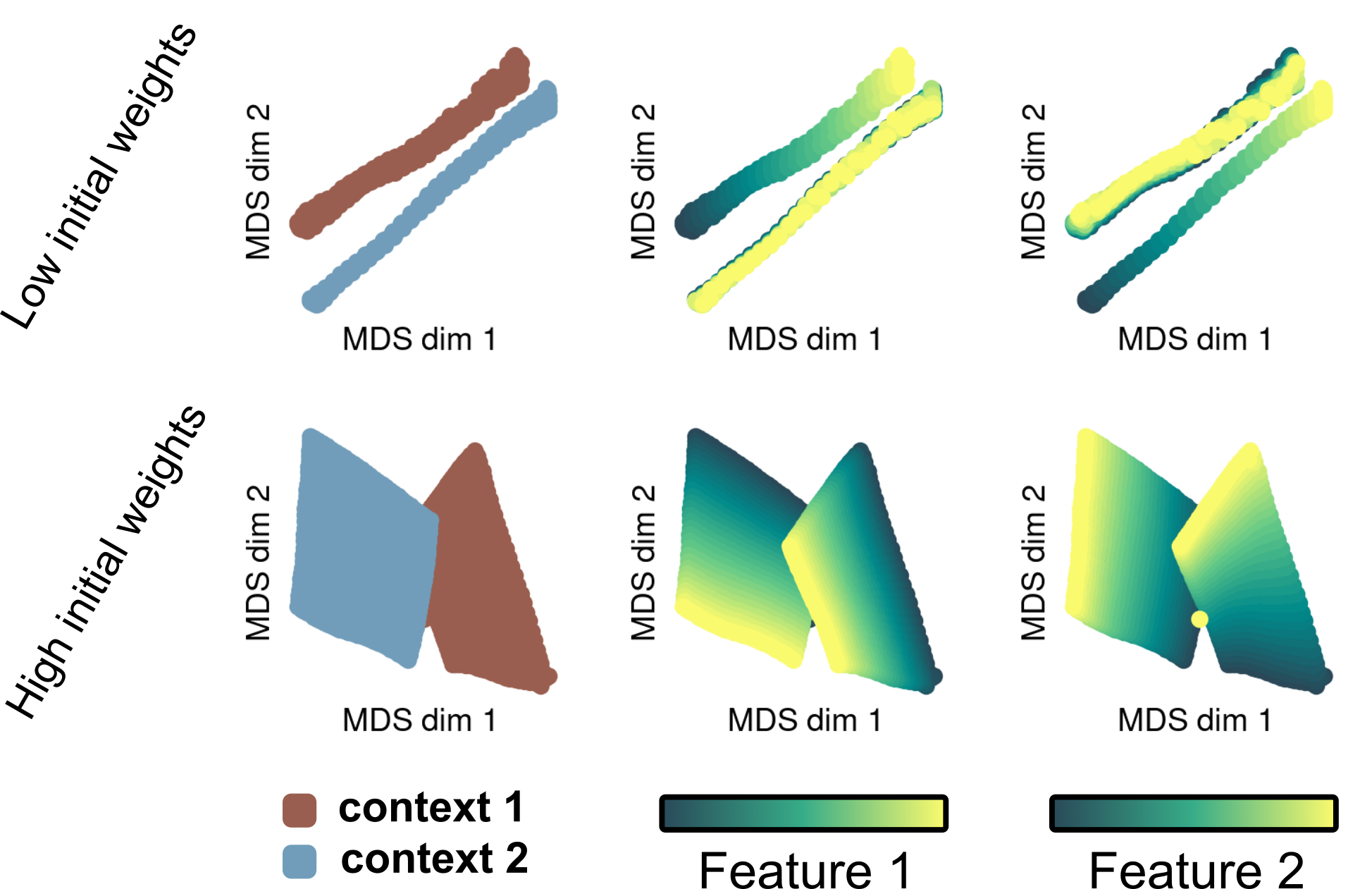}}
\caption{Representational structure in the orthogonal feature task. A multidimensional scaling projection of the representational structure of a 5-layer, tanh, fully connected network with 100 hidden units per layer, trained on the orthogonal feature dataset. At small initial weights the task-irrelevant direction collapses in each context, but remains at large initial weights.}
\label{fig:feature_collapse}
\end{center}
\vskip -0.2in
\end{figure}

The theory's prediction at small initial weights is that for each context the representation will collapse along the feature direction which is irrelevant to the task, as the output label does not change along this direction. At large initial weights the representation will retain the input data structure, i.e. one 2-d square for each context. These two structures at different weight regimes were indeed observed for a two layer ReLU network in \cite{flesch_orthogonal_2022}. Because the prediction here is universal we reran the experiment for a different architecture (5-layer tanh), and found similar results, as shown in \cref{fig:feature_collapse}.

\subsection{MNIST}
\begin{figure}[ht]
\vskip 0.2in
\begin{center}
\centerline{\includegraphics[width=0.8\columnwidth]{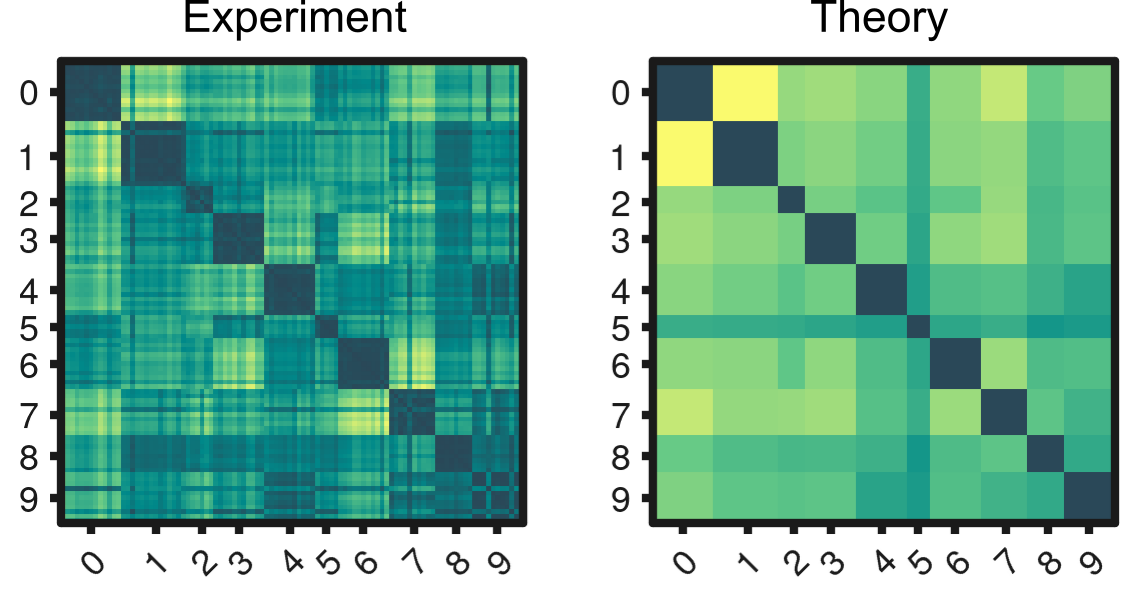}}
\caption{Representational structure on the MNIST dataset at small initial weights. Pairwise distances of the first 100 datapoints in MNIST after averaging over 50 trials of a 4-layer, leaky-ReLU, fully connected network trained on the full dataset (\textbf{left}) compared to the theory after exponential weighing (\textbf{right}). The Pearson correlation coefficient between the theory and experiment is 0.703 and when not factoring in the diagonal blocks it is 0.512.}
\label{fig:mnist_dist}
\end{center}
\vskip -0.2in
\end{figure}

When dealing with more complex datasets we cannot always apply \cref{eq:final_distance} directly. This is because it only predicts an interaction between two datapoints. Even simple geometric constraints such as the triangle inequality are not expected. Nevertheless, in \cref{fig:mnist_dist} we see that part of the representational structure still resembles the theory when a network is trained on the MNIST dataset. Here we applied a simple method to reduce violations of the triangle inequality, as detailed in \cref{sec:exponential_weighing}.

\section{Depth}
In the theory we consider the representation dynamics at ``some intermediate layer $H$". Deep neural networks have many layers at which the representations can be observed. This raises the question: how do the dynamics depend on the depth of the chosen intermediate layer?

\subsection{Predictivity of the effective theory}

\begin{figure}[ht]
\vskip 0.2in
\begin{center}
\centerline{\includegraphics[width=0.8\columnwidth]{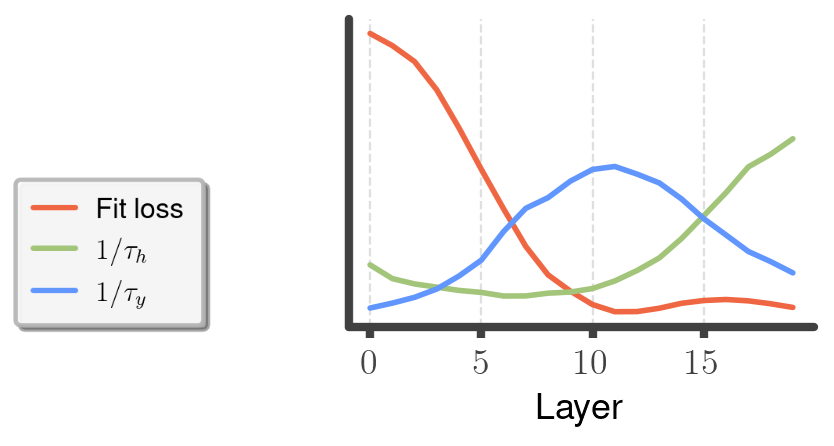}}
\caption{Fit loss and fitted effective learning rates of the theory as a function of the hidden layer depth. The default model (20 fully connected layers, 500 units per layer, leaky ReLU) is trained on the two point dataset, and the considered hidden layer is varied. The y-axis is not to scale. We can see that the theory is accurate only in the second half of the network. In this regime the effective learning rate of the encoder increases with depth, whereas the effective learning rate of the decoder decreases.}
\label{fig:depth}
\end{center}
\vskip -0.2in
\end{figure}

Let us begin by determining at which layers the effective theory is still valid. For the linear approximation to be accurate, we need the representations to start off near each other. Suppose that at small initial weights the average activational gain factor of each layer is some constant $G<1$. The initial representational distance as a function of the depth $n$ scales roughly as
\begin{equation}
    ||dh(0)||^2 \sim G^n ||x_2 - x_1||^2
    .
\end{equation}
This is a decreasing function, so we expect the theory to be more accurate in the later layers of the network. Indeed, as can be seen in \cref{fig:depth} (red) the effective theory models the learning behavior more accurately at the later layers. 

\subsection{Effective Learning Rates}

How do we expect the effective learning rates to change at different hidden layers? The standard gradient descent update contains a sum over the parameters, so it makes changes to a map proportional to the number of parameters it contains. As we consider later hidden layers, the number of parameters in the encoder map increases, whereas the number of parameters in the decoder map decreases. This should result in the effective learning rate for the encoder increasing with the depth and the effective learning rate for the decoder decreasing with the depth. We can see that in \cref{fig:depth} this relationship holds for the later layers of the network, where the theory is accurate, although at earlier layers the effective learning rate for the decoder seems to increase. Notably, these results suggest that representational dynamics in the later layers behave in a similar manner, up to a rescaling of effective learning rates.

\section{Conclusion}

We have introduced an effective theory for the universal part of the learning dynamics common amongst different neural network architectures, and shown it already exhibits structured representational learning, in particular at small initial weights. The aim here is not so much to point to this specific theory as an accurate universal model, rather it is to suggest that some of the essentials required for representation learning may already be contained within gradient descent, as opposed to solely coming from inductive biases contained in the choice of architecture. Therefore, perhaps a more universal perspective to learning dynamics is possible. Additionally, the effective theory underlines the initial weight scale as a crucial factor for the formation of the final representational structure.

A significant limitation to this specific approach for modelling universality is that there is no clear path to extending it to larger datasets. More work is needed to find a satisfying resolution to dealing with more complex data. Additionally, many architectures do have inductive biases that affect the learned representation, and the theory here does not take those into account. The representational effect modelled here likely interacts with these inductive biases, causing potentially interesting effects on the representation that are worth exploring.

\section*{Impact Statement}
This paper presents work whose goal is to increase understanding of deep learning, which may lead to advancements in the field of Machine Learning. There are many potential societal consequences of our work, none which we feel must be specifically highlighted here.

\section*{Acknowledgements}
We thank Erin Grant and Yedi Zhang for useful feedback. This work was supported by a Sir Henry Dale Fellowship from the Wellcome Trust and Royal Society (216386/Z/19/Z) to A.S., and the Sainsbury Wellcome Centre Core Grant from Wellcome (219627/Z/19/Z) and the Gatsby Charitable Foundation (GAT3850).

\bibliography{references}
\bibliographystyle{icml2024}

\newpage
\appendix
\onecolumn

\appendix

\section{Details on the Theoretical Analysis}

Here we provide computational details to the derivation of the effective theory, along with some additional results on the behavior the theory exhibits. 

\subsection{Reduction to a 3-dimensional System}
\label{sec:reduction_to_a_3-dimensional_system}

We begin by modelling our neural network as being represented by a smooth encoder map $h: X \to H$ assigning representations to the inputs and a smooth decoder map $y: H \to Y$ assigning outputs to the representations. To model the two point interaction we restrict to a two datapoint dataset $\mathcal{D} = \{(x_1,y_1),(x_2,y_2)\}$ and take a linear approximation around their representational and output mean:
\begin{equation}
    \begin{split}
        h(x_{i})&=h(\frac{x_{2}+x_{1}}{2})+\frac{1}{2}D_{h}(x_{i}-x_{\neg i})\\ \\
y(x_{i})&=y(h(\frac{x_{2}+x_{1}}{2}))+\frac{1}{2}D_{y}(h(x_{i})-h(x_{\neg i}))\\
&=y(h(\frac{x_{2}+x_{1}}{2}))+\frac{1}{2}D_{y}(D_{h}(x_{i}-x_{\neg i}))
.
    \end{split}
\end{equation}

The mean squared loss in this approximation has the form:
\begin{equation}
    L=\frac{1}{2}\langle||y(h(\frac{x_{2}+x_{1}}{2}))+\frac{1}{2}D_{y}(D_{h}(x_{i}-x_{\neg i}))-y_{i}||^{2}\rangle_{\mathcal{D}}
    .
\end{equation}
Motivated by the assumption of high model expressivity, we apply gradient decent optimization directly with respect to $D_h$, $D_y$ and $\bar{y} := y(h(\frac{x_{2}+x_{1}}{2}))$, resulting in the dynamics:
\begin{equation}
    \begin{split}
        \frac{\mathrm{d}}{\mathrm{d}t} \bar{y} &= -\frac{1}{\tau_{\bar{y}}}\frac{\partial L}{\partial \bar{y}}\\&=-\frac{1}{\tau_{\bar{y}}}\langle\bar{y}+\frac{1}{2}D_{y}(D_{h}(x_{i}-x_{\neg i}))-y_{i}\rangle_{\mathcal{D}}\\&=-\frac{1}{\tau_{\bar{y}}}(\bar{y}-\frac{y_2+y_1}{2})\\
        \\
        \frac{\mathrm{d}}{\mathrm{d}t}  D_h &= -\frac{1}{\tau_h}\frac{\partial L}{\partial D_h}\\&= -\frac{1}{\tau_h} \frac{1}{2}\langle(\frac{1}{2}D_y^\top D_yD_h(x_{i}-x_{\neg i})(x_{i}-x_{\neg i})^\top+D_{y}^\top (\bar{y} -y_{i}) (x_{i}-x_{\neg i})^\top)\rangle_{\mathcal{D}}\\ &= -\frac{1}{\tau_h} \frac{1}{2}D_y^\top(D_y D_h(x_2-x_1)-  (y_2-y_1))(x_2-x_1)^\top\\
        \\
        \frac{\mathrm{d}}{\mathrm{d}t} D_y &= -\frac{1}{\tau_y}\frac{\partial L}{\partial D_y}\\&= -\frac{1}{\tau_y} \frac{1}{2}\langle(\frac{1}{2}D_y D_h(x_i-x_{\neg i})(D_h(x_i-x_{\neg i}))^\top+(\bar{y}-y_{i}) (D_{h}(x_{i}-x_{\neg i}))^\top)\rangle_{\mathcal{D}}\\&=-\frac{1}{\tau_y} \frac{1}{2}(D_y D_h(x_2-x_1)-(y_2-y_1) )(D_{h}(x_2-x_1))^\top
        ,
    \end{split}
\end{equation}
where we used the matrix differentiation identities $\frac{\partial a^\top X b}{\partial X} = a b^\top$, $\frac{\partial a^\top X^\top C X a}{\partial X}=(C+C^\top) X a a^\top$.

The $\bar{y}$ dynamics are decoupled and can be solved directly:
\begin{equation}
    \bar{y}(t) = \frac{y_2+y_1}{2}+(y(0)-\frac{y_2+y_1}{2}) e^{-\frac{t}{\tau_{\bar{y}}}}
    ,
\end{equation}
the solution of which takes the form of exponential decay towards the target output mean.

Define $dh:=D_h(x_2-x_1),dy:=D_y D_h(x_2-x_1), w:=||dy||^2-dy^{\top}(y_2-y_1)$. To simplify the dynamics, we restrict to solutions where the representations only move towards each other or away from each other, i.e.
\begin{equation}
    \frac{\mathrm{d}}{\mathrm{d}t} dh \propto dh
    .
\end{equation}
We find that the derivatives
\begin{equation}
    \begin{split}
        \frac{\mathrm{d}}{\mathrm{d}t} ||dh||^2 &= 2dh^\top \frac{\mathrm{d}}{\mathrm{d}t}dh\\
        &= 2dh^\top \dot{D_h} (x_2-x_1)\\
        &= -\frac{1}{\tau_h} (||dy||^2- dy^\top(y_2-y_1))||x_2-x_1||^2\\
        &= -\frac{1}{\tau_h} ||x_2-x_1||^2 w\\
        \\
        \frac{\mathrm{d}}{\mathrm{d}t} ||dy||^2 &= 2dy^\top \frac{\mathrm{d}}{\mathrm{d}t}dy\\
        &= 2dy^\top (\dot{D_y} D_h (x_2-x_1)+D_y \frac{\mathrm{d}}{\mathrm{d}t} dh)\\
        &= 2dy^\top (\dot{D_y} D_h (x_2-x_1)+\frac{dh^\top \frac{\mathrm{d}}{\mathrm{d}t} dh}{||dh||^2} D_y dh)\\
        &= 2dy^\top (\dot{D_y} D_h (x_2-x_1)+\frac{1}{2}\frac{\frac{\mathrm{d}}{\mathrm{d}t} ||dh||^2}{||dh||^2} D_y dh)\\
        &= 2dy^\top (-\frac{1}{\tau_y} \frac{1}{2}(dy-(y_2-y_1))(dh)^\top dh -\frac{1}{\tau_h}\frac{1}{2} ||x_2-x_1||^2 \frac{w}{||dh||^2} dy)\\
        &= -w(\frac{1}{\tau_y} ||dh||^2 +\frac{1}{\tau_h} ||x_2-x_1||^2 \frac{||dy||^2}{||dh||^2})\\
        \\
        \frac{\mathrm{d}}{\mathrm{d}t} w &= (2dy  - (y_2-y_1))^\top \frac{\mathrm{d}}{\mathrm{d}t}dy\\
        &= (2dy  - (y_2-y_1))^\top (-\frac{1}{\tau_y} \frac{1}{2}(dy-(y_2-y_1))||dh||^2 -\frac{1}{\tau_h}\frac{1}{2} ||x_2-x_1||^2 \frac{w}{||dh||^2} dy)\\
        &= -\frac{1}{\tau_y} \frac{1}{2}(3w-||dy||^2+||y_2-y_1||^2)||dh||^2 -\frac{1}{\tau_h}\frac{1}{2} ||x_2-x_1||^2 \frac{(||dy||^2+w)w}{||dh||^2}
        ,
    \end{split}
\end{equation}
form a self-contained 3 dimensional system:
\begin{equation}
    \begin{split}
        \frac{\mathrm{d}}{\mathrm{d}t} ||dh||^2 &= -\frac{1}{\tau_h} ||x_2-x_1||^2 w\\
        \frac{\mathrm{d}}{\mathrm{d}t} ||dy||^2 &= -w(\frac{1}{\tau_y} ||dh||^2 +\frac{1}{\tau_h} ||x_2-x_1||^2 \frac{||dy||^2}{||dh||^2}) \\
        \frac{\mathrm{d}}{\mathrm{d}t} w &= -\frac{1}{\tau_y} \frac{1}{2}(3w-||dy||^2+||y_2-y_1||^2)||dh||^2 -\frac{1}{\tau_h}\frac{1}{2} ||x_2-x_1||^2 \frac{(||dy||^2+w)w}{||dh||^2}
        .
    \end{split}
    \label{eq:dynamics_2}
\end{equation}
Because this system is self-contained and consists of only 3 scalar variables, it can easily be solved numerically given initial conditions and values for the effective learning rates $1/\tau_h$ and $1/\tau_y$.

\subsection{Expression for the Loss Curve}
\label{sec:expression_for_the_loss_curve}

The loss can be expressed in terms of variables in the 3-dimensional system:
\begin{equation}
    \begin{split}
        L&=\frac{1}{2}\langle||\bar{y}+\frac{1}{2}D_y D_h(x_i-x_{\neg i})-y_{i}||^{2}\rangle_{\mathcal{D}}\\
        &=\frac{1}{2}\langle ||\bar{y} - y_i||^2 +(D_y D_h(x_i-x_{\neg i}))^\top(\bar{y} - y_i) + \frac{1}{4} ||dy||^2\rangle_{\mathcal{D}}\\
        &=\frac{1}{4} (||\bar{y} - y_1||^2+||\bar{y} - y_2||^2 -(dy)^\top(y_2- y_1) + \frac{1}{2} ||dy||^2)\\
        &=\frac{1}{4} (||\bar{y} - y_1||^2+||\bar{y} - y_2||^2 + w - \frac{1}{2} ||dy||^2)\\
        &=\frac{1}{4} (||(y(0)-\frac{y_2+y_1}{2}) e^{-\frac{t}{\tau_{\bar{y}}}}+\frac{y_2-y_1}{2}||^2+||(y(0)-\frac{y_2+y_1}{2}) e^{-\frac{t}{\tau_{\bar{y}}}}-\frac{y_2-y_1}{2}||^2 + w - \frac{1}{2} ||dy||^2)\\
        &=\frac{1}{4} (2||(y(0)-\frac{y_2+y_1}{2}) e^{-\frac{t}{\tau_{\bar{y}}}}||^2 + w + \frac{1}{2} (||y_2-y_1||^2-||dy||^2))
        .
    \end{split}
\end{equation}

\subsection{Final Representational Distance}
\label{sec:final_representational_distance}

The system in \cref{eq:dynamics_2} cannot be solved analytically in the general case, however, we can still gain some analytical insights. For instance, we can determine the final representational distance, which may provide insight into the kinds of representational structures formed. To find this, we exploit the following relationship in the dynamics:
\begin{equation}
    \begin{split}
        \frac{\mathrm{d}}{\mathrm{d}t}\frac{||dy||^2}{||dh||^2} &= \frac{||dh||^2\frac{\mathrm{d}}{\mathrm{d}t}||dy||^2-||dy||^2\frac{\mathrm{d}}{\mathrm{d}t}||dh||^2}{||dh||^4}
        = -\frac{1}{\tau_y}w
        =\frac{\tau_h}{\tau_y} \frac{1}{||x_2-x_1||^2} \frac{\mathrm{d}}{\mathrm{d}t} ||dh||^2
        .
    \end{split}
\end{equation}
Integrating both sides, we can write $||dy||^2$ as a function of $||dh||^2$
\begin{equation}
    ||dy||^2 = \frac{\tau_h}{\tau_y} \frac{1}{||x_2-x_1||^2} ||dh||^4 + \left( \frac{||dy(0)||^2}{||dh(0)||^2} - \frac{\tau_h}{\tau_y} \frac{||dh(0)||^2 }{||x_2-x_1||^2} \right) ||dh||^2
    ,
    \label{eq:y_as_function_of_h}
\end{equation}
reducing the dynamics to a 2-dimensional system:
\begin{equation}
    \begin{split}
        \frac{\mathrm{d}}{\mathrm{d}t} ||dh||^2 &= -\frac{1}{\tau_h} ||x_2-x_1||^2 w\\
        \frac{\mathrm{d}}{\mathrm{d}t} w &= -\frac{1}{2}(-\frac{\tau_h}{{\tau_y}^2} \frac{1}{||x_2-x_1||^2} ||dh||^6 +\frac{1}{\tau_y}||y_2-y_1||^2||dh||^2 + \frac{4}{\tau_y} ||dh||^2 w + \frac{1}{\tau_h}||x_2-x_1||^2\frac{w^2}{||dh||^2} \\&+ \left( \frac{||dy(0)||^2}{||dh(0)||^2} - \frac{\tau_h}{\tau_y} \frac{||dh(0)||^2 }{||x_2-x_1||^2} \right)(\frac{1}{\tau_h}||x_2-x_1||^2 w - \frac{1}{\tau_y}||dh||^4))
        .
    \end{split}
\end{equation}
This system explicitly contains the initial conditions, making it less intuitive to work with. However, it is useful for finding fixed points, as any fixed point of this system must also be one in the original system. To determine what the fixed points are, we set the derivatives to zero.

From the $||dh||^2$ dynamics we immediately find that $w$ is zero at the fixed points:
\begin{equation}
    \frac{\mathrm{d}}{\mathrm{d}t}||dh||^2 =0 \implies w= 0
    ,
\end{equation}
and from the $w$ dynamics we find three potential values for $||dh||^2$:
\begin{equation}
    \begin{split}
        &\frac{\mathrm{d}}{\mathrm{d}t}w=0\\
        \implies& 0=-\frac{\tau_h}{{\tau_y}} \frac{1}{||x_2-x_1||^2} ||dh||^6 -\left( \frac{||dy(0)||^2}{||dh(0)||^2} - \frac{\tau_h}{\tau_y} \frac{||dh(0)||^2 }{||x_2-x_1||^2} \right)||dh||^4 + ||y_2-y_1||^2||dh||^2\\
        \implies& ||dh||^2=0 \text{ or } ||dh||^2 = \frac{1}{2}A_{\text{high}} \pm \sqrt{\frac{1}{4}A^2_{\text{high}} + A^2_{\text{low}}}
        ,
    \end{split}
\end{equation}
where
\begin{equation}
    \begin{split}
        A_{\text{high}} &= \left(\frac{||dh(0)||^2 }{||x_2-x_1||^2} - \frac{\tau_y}{\tau_h} \frac{||dy(0)||^2}{||dh(0)||^2}\right)||x_2-x_1||^2\\
        A_{\text{low}} & =\sqrt{\frac{\tau_y}{\tau_h}} ||x_2-x_1||\cdot ||y_2-y_1||
        .
    \end{split}    
\end{equation}

The fixed point at $||dh||^2 = \frac{1}{2}A_{\text{high}} - \sqrt{\frac{1}{4}A^2_{\text{high}} + A^2_{\text{low}}}$ is negative and is thus not a valid solution. The dynamics will never be able to reach it as $||dh||^2$ cannot go below zero.

The fixed point at $||dh||^2 = \frac{1}{2}A_{\text{high}} + \sqrt{\frac{1}{4}A^2_{\text{high}} + A^2_{\text{low}}}$ has Jacobian
\begin{equation}
    J = \left[\begin{array}{cc}
        0 & -\frac{1}{\tau_{h}}||x_{2}-x_{1}||^{2}\\
        \frac{1}{4}\frac{\tau_{h}}{\tau_{y}^{2}}\frac{1}{||x_{2}-x_{1}||^{2}}(2A_{\text{low}}^{2}+\frac{1}{2}(A_{\text{high}}+\sqrt{A_{\text{high}}^{2}+4A_{\text{low}}^{2}})^{2}) & -\frac{1}{\tau_{y}}(\frac{1}{2}A_{\text{high}}+\sqrt{A_{\text{high}}^{2}+4A_{\text{low}}^{2}})
        \end{array}\right]
        ,
\end{equation}
The trace
\begin{equation}
    \tau = -\frac{1}{\tau_{y}}(\frac{1}{2}A_{\text{high}}+\sqrt{A_{\text{high}}^{2}+4A_{\text{low}}^{2}})
\end{equation}
is negative and the determinant
\begin{equation}
    D = \frac{1}{4}\frac{1}{\tau_{y}^{2}}(2A_{\text{low}}^{2}+\frac{1}{2}(A_{\text{high}}+\sqrt{A_{\text{high}}^{2}+4A_{\text{low}}^{2}})^{2})
\end{equation}
is positive. This fixed point is therefore always stable.

The fixed point at $||dh||^2=0$ has Jacobian
\begin{equation}
    J=\left[\begin{array}{cc}
        0 & -\frac{1}{\tau_{h}}||x_{2}-x_{1}||^{2}\\
        \frac{1}{2}(\frac{1}{\tau_{h}}||x_{2}-x_{1}||^{2}\frac{w^{2}}{||dh||^{4}}-\frac{1}{\tau_{y}}||y_{2}-y_{1}||^{2}) & (\frac{1}{2}\frac{1}{\tau_{y}}A_{\text{high}}-\frac{1}{\tau_{h}}||x_{2}-x_{1}||^{2}\frac{w}{||dh||^{2}})
        \end{array}\right]
        .
\end{equation}
This Jacobian cannot be directly evaluated at $w=0,||dh||^2=0$ because of the undetermined term $\frac{w}{||dh||^2}$. By replacing $\frac{w}{||dh||^2}$ with the direction of approach $\frac{b}{a}$ we can solve for eigenvectors
\begin{equation}
    \left[\begin{array}{cc}
0 & -\frac{1}{\tau_{h}}||x_{2}-x_{1}||^{2}\\
\frac{1}{2}(\frac{1}{\tau_{h}}||x_{2}-x_{1}||^{2}\frac{b^{2}}{a^{2}}-\frac{1}{\tau_{y}}||y_{2}-y_{1}||^{2}) & (\frac{1}{2}\frac{1}{\tau_{y}}A_{\text{high}}-\frac{1}{\tau_{h}}||x_{2}-x_{1}||^{2}\frac{b}{a})
\end{array}\right]\left[\begin{array}{c}
a\\
b
\end{array}\right]=\lambda \left[\begin{array}{c}
 a\\
 b
\end{array}\right]
\end{equation}
to find
\begin{equation}
    v_{\pm}=\left[\begin{array}{c}
1\\
-\frac{\tau_{h}}{\tau_{y}}\frac{A_{\text{high}}\pm\sqrt{A_{\text{high}}^{2}+4A_{\text{low}}^{2}}}{2||x_{2}-x_{1}||^{2}}
\end{array}\right]
\end{equation}
with one positive and one negative eigenvalue
\begin{equation}
    \lambda_{\pm}=\frac{1}{\tau_{y}}\frac{A_{\text{high}}\pm\sqrt{A_{\text{high}}^{2}+4A_{\text{low}}^{2}}}{2}
    .
\end{equation}
There is always one direction along which perturbations increase, so this fixed point is not stable. In the limit of infinite time we expect the final representational distance to go to the only stable fixed point:
\begin{equation}
    ||dh(\infty)||^2 = \frac{1}{2}A_{\text{high}} + \sqrt{\frac{1}{4}A^2_{\text{high}} + A^2_{\text{low}}}
    .
\end{equation}

\subsection{Solution for Identical Outputs}
\label{sec:solution_for_identical_outputs}
In the case that the target outputs of the two datapoints are identical, i.e. $y_1 = y_2$, an exact solution can be found. In this case we have by definition $w=||dy||^2$, reducing the system to:
\begin{equation}
    \begin{split}
        \frac{\mathrm{d}}{\mathrm{d}t}||dh||^{2}	&=-\frac{1}{\tau_h} ||x_2-x_1||^2 ||dy||^{2}\\
\frac{\mathrm{d}}{\mathrm{d}t} ||dy||^2 &= -||dy||^2(\frac{1}{\tau_y} ||dh||^2 +\frac{1}{\tau_h} ||x_2-x_1||^2 \frac{||dy||^2}{||dh||^2}) 
    \end{split}
\end{equation}
Using \cref{eq:y_as_function_of_h} we can write self-contained $||dh||^2$ dynamics:
\begin{equation}
    \frac{\mathrm{d}}{\mathrm{d}t}||dh||^2 = -\frac{1}{\tau_y} ||dh||^4 +\frac{1}{\tau_y} A_\text{high} ||dh||^2
    ,
\end{equation}
which is a Bernoulli equation and has solution
\begin{equation}
    ||dh(t)||^2 = \frac{A_\text{high}}{1 + (\frac{A_\text{high}}{||dh(0)||^2}-1)e^{-A_\text{high} t}}
    .
\end{equation}
Plugging this in \cref{eq:y_as_function_of_h} gives us the $||dy||^2$ solution:
\begin{equation}
    ||dy||^2 = \frac{\tau_h}{\tau_y} \frac{A_\text{high}^2}{||x_2-x_1||^2} \frac{(1-\frac{A_\text{high}}{||dh(0)||^2})e^{-A_\text{high} t}}{(1 + (\frac{A_\text{high}}{||dh(0)||^2}-1)e^{-A_\text{high} t})^2}
    .
\end{equation}

\subsection{Lazy Learning Dynamics}
\label{sec:lazy_learning_dynamics}

At large weights $w \approx ||dy||^2$, so increasing the weights scales the initial values as \begin{equation}
    \begin{split}
        ||dh||^{2}&\to G||dh||^{2}\\
        ||dy||^{2}&\to G^{2}||dy||^{2}\\\
        w&\to G^{2}w
    \end{split}
    .
\end{equation}
Under the variable change
\begin{equation}
    \begin{split}
        {||dh||^{2}}'	&=G||dh||^{2}\\
        {||dy||^{2}}'	&=G^{2}||dy||^{2}\\
        w' &=G^{2}w\\
        t' &=G^{-1}t
    \end{split}
    ,
\end{equation}
the dynamics change to 
\begin{equation}
    \begin{split}
        \frac{\mathrm{d}}{\mathrm{d}t'} {||dh||^2}' &= -\frac{1}{\tau_h} ||x_2-x_1||^2 w'\\
        \frac{\mathrm{d}}{\mathrm{d}t'} {||dy||^2}' &= -w'(\frac{1}{\tau_y} {||dh||^2}' +\frac{1}{\tau_h} ||x_2-x_1||^2 \frac{{||dy||^2}'}{{||dh||^2}'}) \\
        \frac{\mathrm{d}}{\mathrm{d}t'} w' &= -\frac{1}{\tau_y} \frac{1}{2}(3w'-{||dy||^2}'+G^{-2} ||y_2-y_1||^2){||dh||^2}' -\frac{1}{\tau_h}\frac{1}{2} ||x_2-x_1||^2 \frac{({||dy||^2}'+w')w'}{{||dh||^2}'}
        .
    \end{split}
\end{equation}

Taking the large initial weight limit $G \to \infty$, these dynamics reduce to the dynamics for identical outputs. We can thus expect a similar solution for large initial weights, as the one derived in \cref{sec:solution_for_identical_outputs}, i.e. smooth exponential decay. The training loss will therefore also exponentially decay to zero without a plateau at large initial weights, which is similar what is observed in different neural networks.

\subsection{Random Feature Learning}
\label{sec:random_feature_learning}

A characteristic of the lazy learning regime is that it learns random features, as opposed to the more consistent structured features learned at small initialization. We can explicitly demonstrate the vanishing of randomness in the representation at small initial weights, by assuming the initial weights are Gaussian distributed, such that $\frac{||dh(0)||^{2}}{||x_{i}-x_{\neg i}||^{2}}\sim\mathcal{N}(G,G^{2}),\frac{||dy(0)||^{2}}{||dh(0)||^{2}}\sim\mathcal{N}(G,G^{2})$, where $G$ is some overall gain factor at initialization. The exact scaling of these distributions with the initial gain is not important here, this is merely done for illustrative purposes. We can write
\begin{equation}
    \begin{split}
        P(||dh(\infty)||^{2}<h)&=P(\frac{1}{2}A_{\text{high}}+\sqrt{\frac{1}{4}A_{\text{high}}^{2}+A_{\text{low}}^{2}}<h)\\
        &=P(\sqrt{A_{\text{high}}^{2}+4A_{\text{low}}^{2}}<2h-A_{\text{high}})\\
        &=P(A_{\text{high}}^{2}+4A_{\text{low}}^{2}<(2h-A_{\text{high}})^2\text{ and }2h-A_{\text{high}}>0)\\
        &=P(4A_{\text{low}}^{2}<4h^2-4 h A_{\text{high}} \text{ and }2h-A_{\text{high}}>0)\\
        &=P(A_{\text{high}}h<h^2-A_{\text{low}}^{2}\text{ and }A_{\text{high}}<2h)\\
    \end{split}
    ,
\end{equation}
for positive $h$.
Since $-A_\text{low}^2 < h^2$, we have $h^2 - A_\text{low}^2 <2h^2$, so in the event that $A_\text{high} h < h^2 - A_\text{low}^2$, the equality $A_\text{high} < 2h$ is automatically satisfied. Thus,
\begin{equation}
    \begin{split}
        P(||dh(\infty)||^{2}<h)
       &=P(A_{\text{high}}h<h^{2}-A_{\text{low}}^{2})\\
    \end{split}
    .
\end{equation}
The distribution on $A_\text{high}$ is Gaussian:
\begin{equation}
\begin{split}
    A_\text{high}&=||x_2-x_1||^{2}(\frac{||dh(0)||^{2}}{||x_2-x_1||^{2}}-\frac{||dy(0)||^{2}}{||dh(0)||^{2}})\\
    &\sim\mathcal{N}(||x_2-x_1||^{2}G-||x_2-x_1||^{2}G,||x_2-x_1||^{4}G^{2}+||x_2-x_1||^{4}G^{2})\\
    &=\mathcal{N}(0,2||x_2-x_1||^{4}G^{2})
\end{split}
\end{equation}
such that
\begin{equation}
    \begin{split}
        P(||dh(\infty)||^{2}<h)
       &=\Phi(\frac{h-\frac{A_{\text{low}}^{2}}{h}}{\sqrt{2}||x_2-x_1||^{2}G})
    \end{split}
    ,
\end{equation}
giving us a final representational distribution:
\begin{equation}
    \begin{split}
        f(h) &= \frac{\mathrm{d}}{\mathrm{d}h}\Phi(\frac{h-\frac{A_{\text{low}}^{2}}{h}}{\sqrt{2}||x_2-x_1||^{2}G})\\
        &=\frac{1+\frac{A_{\text{low}}^{2}}{h^{2}}}{\sqrt{2}||x_2-x_1||^{2}G}\phi(\frac{h-\frac{A_{\text{low}}^{2}}{h}}{\sqrt{2}||x_2-x_1||^{2}G})\\
        &=\frac{1+\frac{A_{\text{low}}^{2}}{h^{2}}}{2\sqrt{\pi}||x_2-x_1||^{2}G}e^{-\frac{1}{2}(\frac{h-\frac{A_{\text{low}}^{2}}{h}}{\sqrt{2}||x_2-x_1||^{2}G})^{2}}
    \end{split}
    ,
\end{equation}
We can see that as we take the limit of small initial weights $G \to 0$, the distribution converges, removing any randomness in the final structure.

\section{Experimental Details}
\label{sec:experiment_details}
Here we provide additional details and methods used in the experiments. For all experiments we used the open-source library PyTorch. We chose stochastic gradient descent as an optimizer, as it was used for the theory derivation. The hidden layer $H$ considered for all models is always the layer exactly halfway through the network.

\subsection{Two Point Experiments}
\label{sec:two_point_experiments}

\paragraph{Dataset.}
The dataset used here is composed of two datapoints with 1-dimensional inputs and outputs. The inputs are $-1$ and $-1 + dx$, the outputs are $0.6$ and $0.6 + dy$ respectively, where $dx$ has been set to 0.5 and $dy$ set to 1.

\paragraph{Architectures.}
The architectures used are all variants of deep neural networks where each layer has the same hidden dimension, all layers fully connected unless specified otherwise. All models are initialized using the Xavier normal initialization with gain parameter chosen to display rich learning behavior. Each layer has biases and these are initialized at zero. Learning rates are chosen to produce smooth loss curves while still converging within the 6000 epochs. The different hyperparameters can be found in \cref{tab:hyper}. For the convolution architecture the connections between hidden layers are replaced with convolution layers, each with 20 channels and kernel size 11. The skip connection architecture has a skip connection on each layer jumping over the next two layers. For the dropout network a dropout probability of 0.1 was used on all units.

\begin{table}[t]
\caption{Hyperparameters of the different architectures used in 2 point experiments.}
\label{tab:hyper}
\vskip 0.15in
\begin{center}
\begin{small}
\begin{sc}
\begin{tabular}{lcccccr}
\toprule
Experiment & nonlinearity & init gain & learning rate & epochs & units & hidden layers\\
\midrule
Default    & leaky relu & 1.0 & 0.005 & 6000 & 500 & 20\\
Convolution & leaky relu & 0.95 & 0.002 & 6000 & 500 & 10\\
Skip connections    & leaky relu & 0.1 & 0.01 & 6000 & 500 & 20\\
Shallow    & leaky relu & 0.35 & 0.01 & 6000 & 500 & 2\\
Narrow     & leaky relu & 1.05 & 0.005 & 6000 & 50 & 20\\
Dropout      & leaky relu & 1.0 & 0.005 & 6000 & 500 & 10\\
Linear      & none & 0.85 & 0.00125 & 6000 & 500 & 20\\
ReLU   & relu & 1.0 & 0.005 & 6000 & 500 & 20\\
ELU   & elu & 0.95 & 0.00045 & 6000 & 500 & 20\\
Tanh   & tanh & 0.85 & 0.00125 & 6000 & 500 & 20\\
Swish   & swish & 1.7 & 0.0025 & 6000 & 500 & 20\\
\midrule
Low initial weights    & leaky relu & 0.9 & 0.015 & 6000 & 500 & 20\\
Intermediate initial weights    & leaky relu & 1.1 & 0.0025 & 6000 & 500 & 20\\
High initial weights    & leaky relu & 1.3 & 0.00075 & 6000 & 500 & 20\\
Very high initial weights    & leaky relu & 1.4 & 0.00025 & 6000 & 500 & 20\\
\bottomrule
\end{tabular}
\end{sc}
\end{small}
\end{center}
\vskip -0.1in
\end{table}

\paragraph{Fitting Procedure.}
To determine the values of $1/\tau_h$ and $1/\tau_y$ we optimize the squared fit loss:
\begin{equation}
    \int_\text{epochs} (||dh(t)||^2 - ||dh(t)||^2_\text{theory})^2 dt + \int_\text{epochs} (||dy(t)||^2 - ||dy(t)||^2_\text{theory})^2 dt + \int_\text{epochs} (w(t) - w(t)_\text{theory})^2 dt
    ,
    \label{eq:fit_loss}
\end{equation}
where $||dh(t)||^2_\text{theory}$, $||dy(t)||^2_\text{theory}$ and $w(t)_\text{theory}$ are the numerical solutions to \cref{eq:dynamics} for the choice of $1/\tau_h$ and $1/\tau_y$. Additionally, for the loss curves, the parameter $1/\tau_{\bar{y}}$ was fit in the same manner.

\paragraph{Final Representational Distance.}
For the comparison of the final representational distance to the theory the same hyperparameter settings were used. First, for each architecture, effective learning rates were fit and averaged over 50 trials at fixed input and output distances $dx=0.5,dy=1$. Next, we ran 5*5 trials for each architecture with varying input distance $dx$ between 0.5 and 1.5, and varying output distance $dy$ between 0 and 1, comparing them to the theory prediction from the averaged effective learning rates. Trials which did not converge were discarded.

\subsection{XOR Task}
\label{sec:xor_task}

\paragraph{Dataset.}
The dataset used has 4 datapoints, 2 input dimensions and 1 output dimension. Explicitly, the 4 datapoints are $\{x: (0,0), y:0\}$, $\{x: (1,0), y:1\}$, $\{x: (0,1), y:1\}$ and $\{x: (1,1), y:0\}$.

\paragraph{Architecture.}
The architecture used is the same as the default architecture in the two point experiments. The hyperparameters used can be found in \cref{tab:hyper_toy} (top). For the theory prediction, the effective learning rates from the two point experiments fitted on the default architecture were used, as it also has 20 layers, 500 units and leaky ReLU.

\begin{table}[t]
\caption{Hyperparameters used in the experiments on the different toy datasets.}
\label{tab:hyper_toy}
\vskip 0.15in
\begin{center}
\begin{small}
\begin{sc}
\begin{tabular}{lcccccr}
\toprule
Experiment & nonlinearity & init gain & learning rate & epochs & units & hidden layers\\
\midrule
XOR Low weights    & leaky relu & 0.7 & 0.015 & 1000 & 500 & 20\\
XOR Intermediate weights    & leaky relu & 1.25 & 0.002 & 1000 & 500 & 20\\
XOR High weights    & leaky relu & 2.0 & 0.000000003 & 1000 & 500 & 20\\
XOR 2 layers    & leaky relu & 0.0015 & 0.03 & 40000 & 500 & 1\\
\midrule
Blobs Very low weights    & tanh & 0.275 & 0.05 & 2000 & 100 & 4\\
Blobs Low weights    & tanh & 0.5 & 0.015 & 1500 & 100 & 4\\
Blobs High weights    & tanh & 2.0 & 0.0015 & 750 & 100 & 4\\
\midrule
MINST Trajectories & leaky relu & 0.5 & 0.003 & 200 & 200 & 4\\
MINST Structure & leaky relu & 0.5 & 0.005 & 200 & 200 & 4\\
\bottomrule
\end{tabular}
\end{sc}
\end{small}
\end{center}
\vskip -0.1in
\end{table}

\subsection{Feature Collapse Experiment}

\label{sec:feature_collapse_experiment}

\paragraph{Dataset.} The dataset is similar to one used in \cite{flesch_orthogonal_2022}, but with an increased number datapoints, as this is required to preserve the topology of the representation due to the added depth in the architecture. The 2*30*30 datapoints are 5*5 pixel images of Gaussian blobs, with varying x and y position of the mean. The Gaussian variance is kept constant. Two additional input dimensions are added for a one-hot encoding of the context variable, resulting in 27 dimensional input data. The output data has a single dimension and is identical the x position for the half of the dataset in the first context, and the y position for the second half. 

\paragraph{Architecture.}

The architecture used is a 5 layer fully connected deep neural network with tanh nonlinearity. Hyperparameters can be found in \cref{tab:hyper_toy} (middle).

\subsection{Experiment on MNIST}

\label{sec:experiment_on_mnist}

\paragraph{Dataset.}
The full training set of MNIST handwritten digit classification was used during training. The representational structure was plotted for the first 100 digits, sorted by digit. For the learning dynamics, a zero digit and a one digit were selected at random. The digits zero and one were chosen since they look fairly distinct, compared to e.g. three and eight. Trajectories look slightly worse for more similar digits, but still roughly share a similar shape to the effective theory.

\paragraph{Architecture.}
The architecture used is a fully connected, leaky ReLU network, with only 4 hidden layers instead of the usual 20 to allow for faster convergence in the feature learning regime. The hyperparameters used for both the learning dynamics and representational structure experiments can be found in \cref{tab:hyper_toy} (bottom). Initial weight parameters were chosen to display rich learning behaviour, while still converging within reasonable time.

\paragraph{Representational Structure}
For comparing the final representational structure to the theory prediction, the model was trained 50 times at varying random seeds. The resulting representational distances between the first 100 digits in the dataset were averaged. The Pearson correlation coefficient is computed between the vector of all pairwise final representational distances and their respective theory predictions from \cref{eq:final_distance}.

\paragraph{Exponential Weighing}

\label{sec:exponential_weighing}

Applying \cref{eq:final_distance} directly makes little sense, as a prediction for pairwise representational distances does not necessarily directly translate into a prediction for the full dataset representation. Even when assuming pairwise distances in the full dataset behave independently, the two datapoint prediction may end up breaking the triangle inequality, i.e. it may predict that representations of datapoints A and B are close, as well as B and C, but still predict A and C to be far away. Thus, we must apply a preprocessing step to take the geometry into account.

One somewhat arbitrary choice of still making predictions is by using exponential weighing, i.e. adjust the representational distance predictions based on nearby datapoints
\begin{equation}
     ||dh_{i,j}(\infty)||^2_\text{pred} = \sum_{k,l} e^{-||dh_{i,k}(\infty)||^2_\text{theory}} e^{-||dh_{l,j}(\infty)||^2_\text{theory}} ||dh_{k,l}(\infty)||^2_\text{theory}
     .
\end{equation}
In this case the large distance between A and C would cause the predicted distance between A and B, and B and C to be large too. Thus, the method reduces issues with the triangle inequality. Although simplistic and without theoretical guarantees, the method is easy to vectorize and can thus be used on a large part of the dataset.

\section{Supplemental Figures}

\subsection{Alternative Dynamics}
\label{sec:alternative_dynamics}

When comparing the effective theory's dynamics to that of the neural networks, the two effective learning rates need to be fit. Given the simplicity of the learning dynamics, one may wonder if fitting these two parameters will lead to spurious conclusions about the accuracy of the effective theory as a model, since a system with enough freedom may be able to fit any simple dynamics.

To demonstrate that the specific form of \cref{eq:dynamics} is important to be able to fit the dynamics well, consider four slightly altered versions of this system, where we added a factor two, a square, removed a minus sign, and remove a term respectively.

\begin{equation}
    \begin{split}
        \frac{\mathrm{d}}{\mathrm{d}t} ||dh||^2 =& -\frac{1}{\tau_h} ||x_2-x_1||^2 w^{\textcolor[HTML]{0358f8}{\textbf{2}}}\\
        \frac{\mathrm{d}}{\mathrm{d}t} ||dy||^2 =& - \textcolor[HTML]{9f3e5f}{\textbf{2}} w(\frac{1}{\tau_y} ||dh||^2 +\textcolor[HTML]{698e3d}{\mathbf {\frac{1}{\tau_h} ||x_2-x_1||^2 \frac{||dy||^2}{||dh||^2} }} ) \\
        \frac{\mathrm{d}}{\mathrm{d}t} w =& -\frac{1}{\tau_y} \frac{1}{2}(3w-||dy||^2+||y_2-y_1||^2)||dh||^2 \textcolor[HTML]{fe9d35}{\textbf{+}} \frac{1}{\tau_h}\frac{1}{2} ||x_2-x_1||^2 \frac{(||dy||^2+w)w}{||dh||^2}
        ,
    \end{split}
    \label{eq:alternative_dynamics}
\end{equation}

When fitting these altered dynamics to the default architecture using the same procedure as before, we find that the squared fit loss (\cref{eq:fit_loss}) increases from 0.92 to 15.35, 16.61, 2.99 and 5.64 respectively.

\begin{figure}[ht]
\vskip 0.2in
\begin{center}
\centerline{\includegraphics[width=\columnwidth]{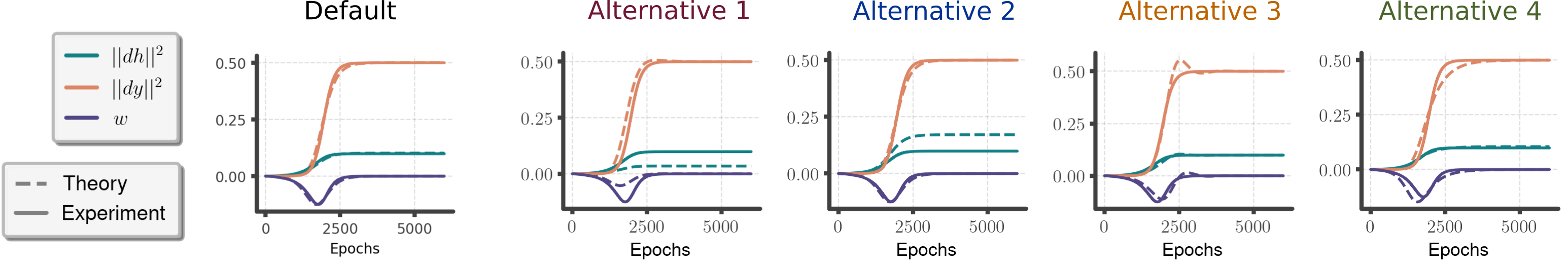}}
\caption{Comparisons of the effective theory (\cref{eq:dynamics}) and four alternative dynamics (\cref{eq:alternative_dynamics}) to the training dynamics of the default architecture (20 fully connected layers, 500 units per layer, leaky ReLU) on two datapoints. Trajectories of the alternatives are not as good of a fit compared to the effective theory.}
\label{fig:trajectories_alternatives}
\end{center}
\vskip -0.2in
\end{figure}

\subsection{Loss Curves for Varying Architectures}

\begin{figure}[ht]
\vskip 0.2in
\begin{center}
\centerline{\includegraphics[width=\columnwidth]{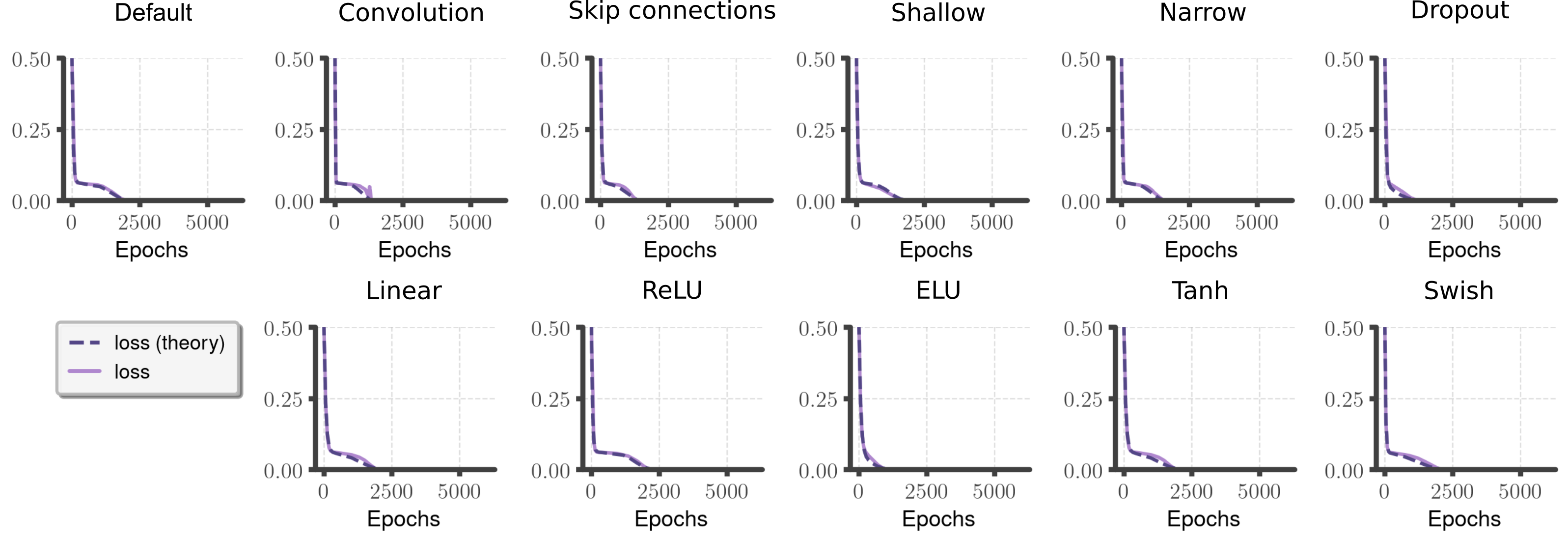}}
\caption{Training loss amongst different architectures. Training loss on the two point experiment compared to the prediction from the effective theory. The first two effective learning rates fitted in the 3-dimensional system were re-used and the final one was fitted using the same procedure.}
\label{fig:loss_among_archtectures}
\end{center}
\vskip -0.2in
\end{figure}

Just as with the dynamics of the 3-dimensional system, we can fit the loss expression from the theory \cref{eq:loss_theory} to the training loss, amongst varying architectures, as is shown in \cref{fig:loss_among_archtectures}. Using the already fitted dynamics of the 3-dimensional system, this requires the fitting of one more unknown parameter, namely $1/\tau_{\bar{y}}$. 

\subsection{Very Low Weight Limit in the Orthogonal Feature Task}

\begin{figure}[ht]
\vskip 0.2in
\begin{center}
\centerline{\includegraphics[width=0.6\columnwidth]{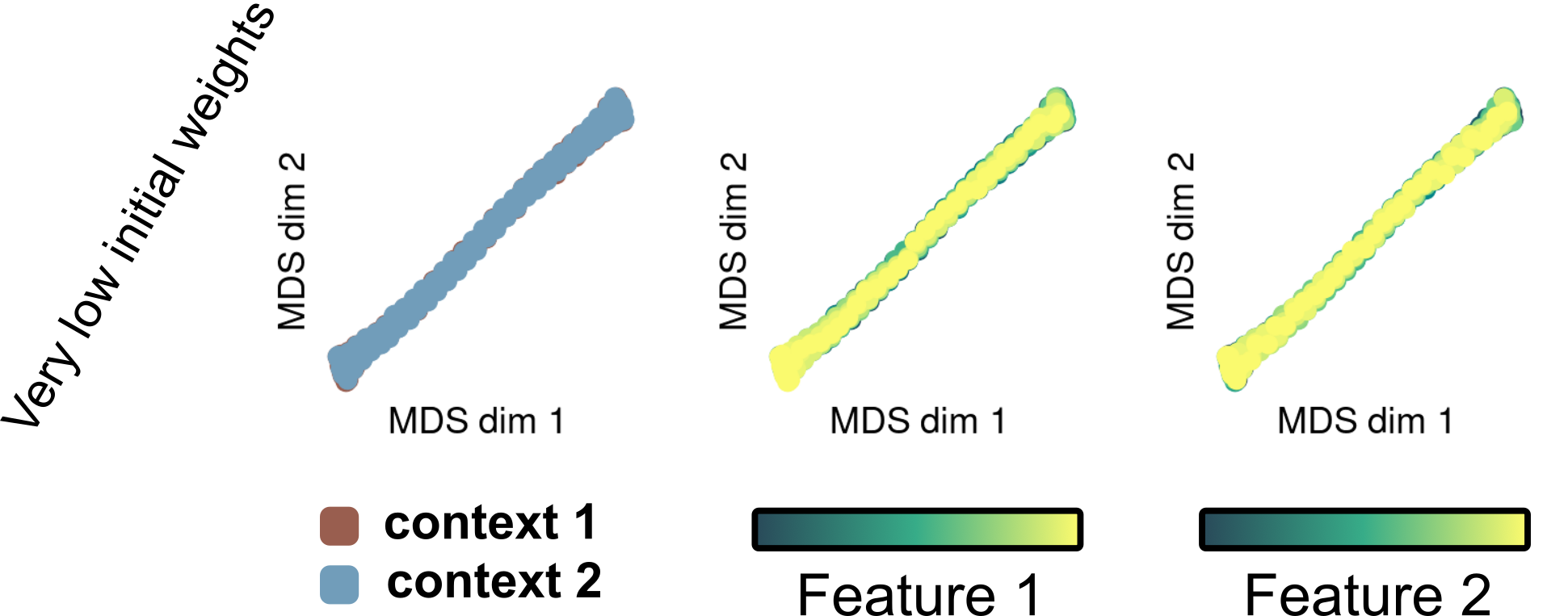}}
\caption{Representational structure in the orthogonal feature task at very low initial weights. A multidimensional scaling projection of the representational structure of a 4-layer, tanh, fully connected network with 100
hidden units per layer, trained on the orthogonal feature dataset. At very small initial weights the representational structure collapses to a single line.}
\label{fig:feature_collapse_vlow}
\end{center}
\vskip -0.2in
\end{figure}

One could argue that for the orthogonal feature task at low initial weights the representation should collapse to a single line, not two disjoint lines. This is because for every point in each context there is a respective point with the same output in the other context, meaning their output difference is zero. Therefore, by \cref{eq:final_distance} their representational distance should also end up at zero. Yet, we still find two separate lines in the representation. This is because the difference in inputs here is much larger than for points along the irrelevant feature, so their representational distance will take longer to go to zero in the small weight limit. Indeed, if we initialize even smaller, we find that the representation collapses to a single line, as can be seen in \cref{fig:feature_collapse_vlow}.

\subsection{Validation of the Linear Approximation}

\begin{figure}[ht]
\vskip 0.2in
\begin{center}
\centerline{\includegraphics[width=0.65\columnwidth]{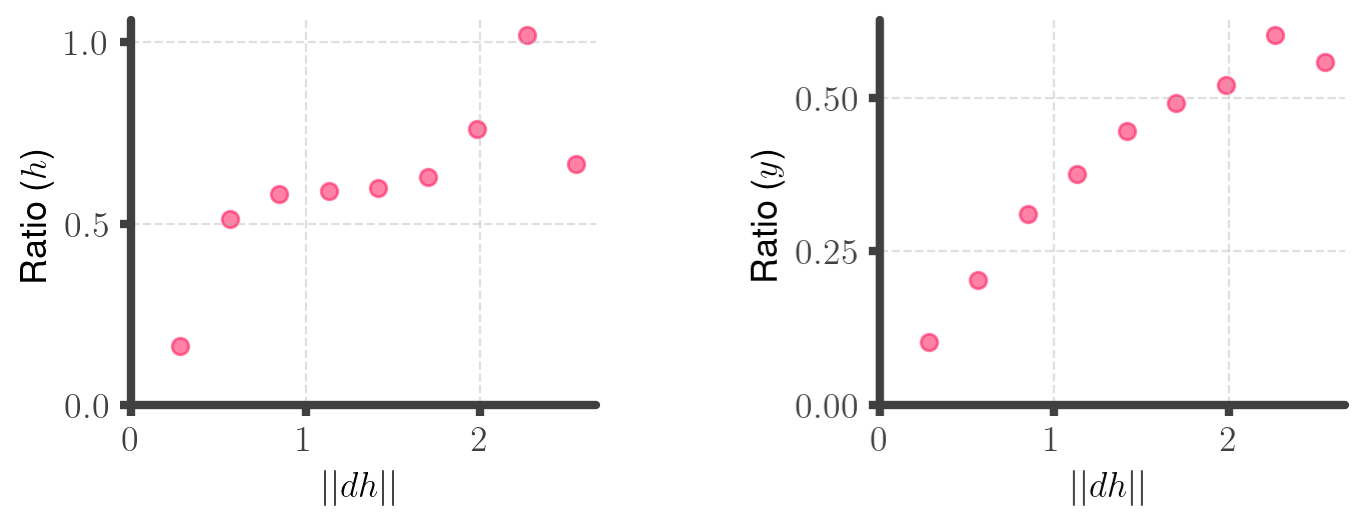}}
\caption{The ratio between the first order term and the second order term in the Taylor expansions of $h$ (\textbf{left}) and $y$ (\textbf{right}) respectively for pairs of datapoints in MNIST during training. We randomly selected 1000 pairs of datapoints, and then grouped them into 10 bins based on their representational distance.  These results are shown after training the MNIST architecture from \cref{tab:hyper_toy} for 30 epochs. We can see a trend for both maps where the linear term dominates for nearby datapoints but gets worse the further they are away. This trend is not present at initialization but does appear very quickly, becoming visible even at a single epoch.}
\label{fig:MNIST_approximation}
\end{center}
\vskip -0.2in
\end{figure}

One may wonder to what extend the linear approximation taken in \cref{eq:linear_approximation} actually holds during training. For the effective theory to accurately model interactions, we would like the first order term to dominate for pairs of nearby datapoints. One way of validating the linear approximation is to compare the first order Taylor term to higher terms. We see in \cref{fig:MNIST_approximation} that it holds up reasonably well for close enough pairs of datapoints.

\subsection{Adam Optimizer}

\begin{figure}[ht]
\vskip 0.2in
\begin{center}
\centerline{\includegraphics[width=0.4\columnwidth]{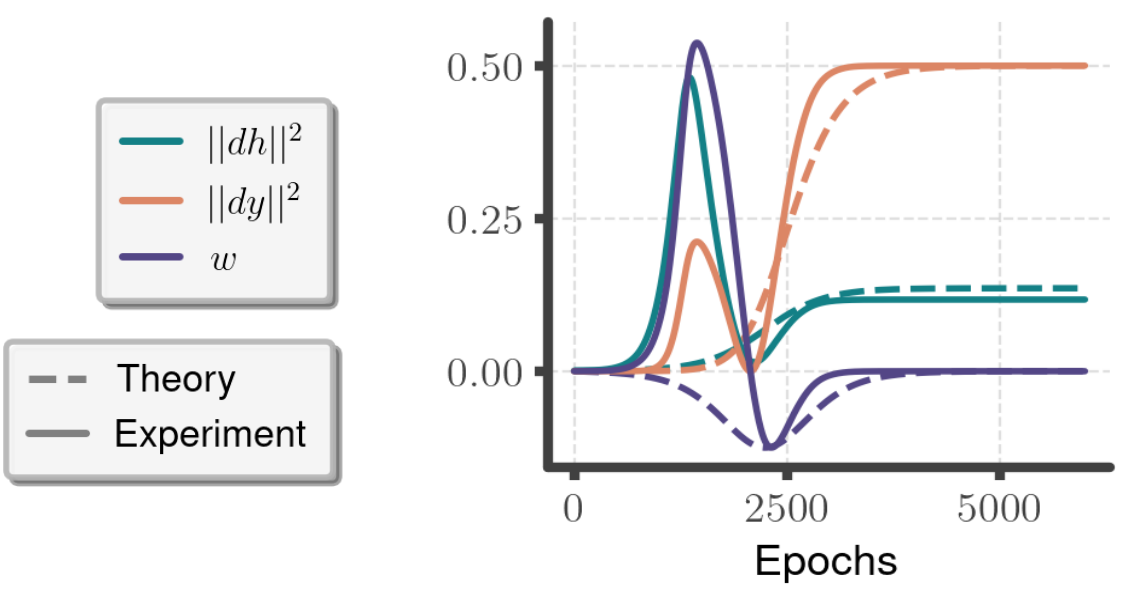}}
\caption{Learning dynamics of the default architecture (20 layers, 500 units, leaky ReLU) trained on two data points using with Adam, compared against the theory after fitting two constants. The fit is not very good, especially in the earlier part of the dynamics.}
\label{fig:adam_trajectories}
\end{center}
\vskip -0.2in
\end{figure}

The dynamics in \cref{eq:dynamics} were derived on the assumption that the neural network is trained using gradient descent. In practice, other optimization schemes are used to train networks, such as Adam. It is a priori not clear that the derived results translate to different optimizers. Indeed, as can be seen in \cref{fig:adam_trajectories} the effective theory does not fit the dynamics of a neural network trained using Adam very well.

\begin{figure*}
  \includegraphics[width=0.9\textwidth]{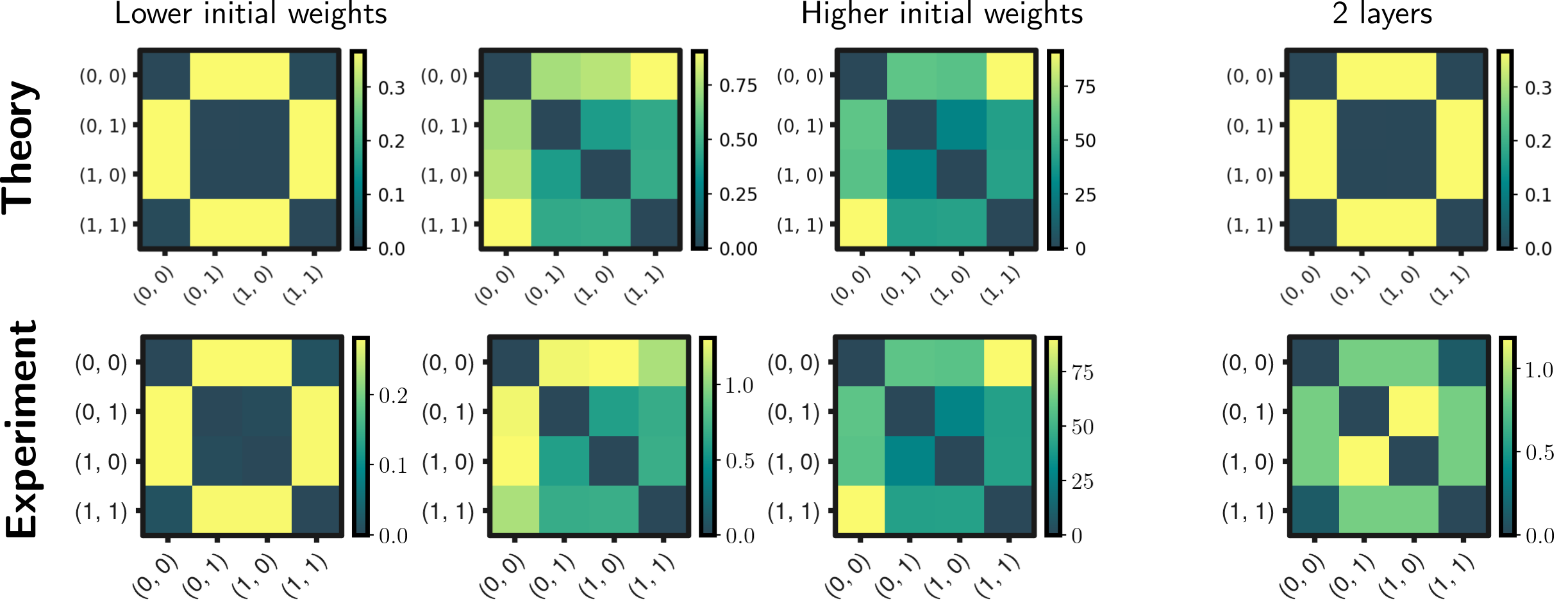}
  \caption{Representational structure in the XOR task, with Adam used as an optimizer. The learned hidden representational pair-wise distances match the theory prediction at varying initial weight scales for the default 20 layer network (\textbf{left}), but not for a 2 layer network at small weights  (\textbf{right}).}
  \label{fig:adam_XOR}
\end{figure*}

\begin{figure}[ht]
\vskip 0.2in
\begin{center}
\centerline{\includegraphics[width=0.6\columnwidth]{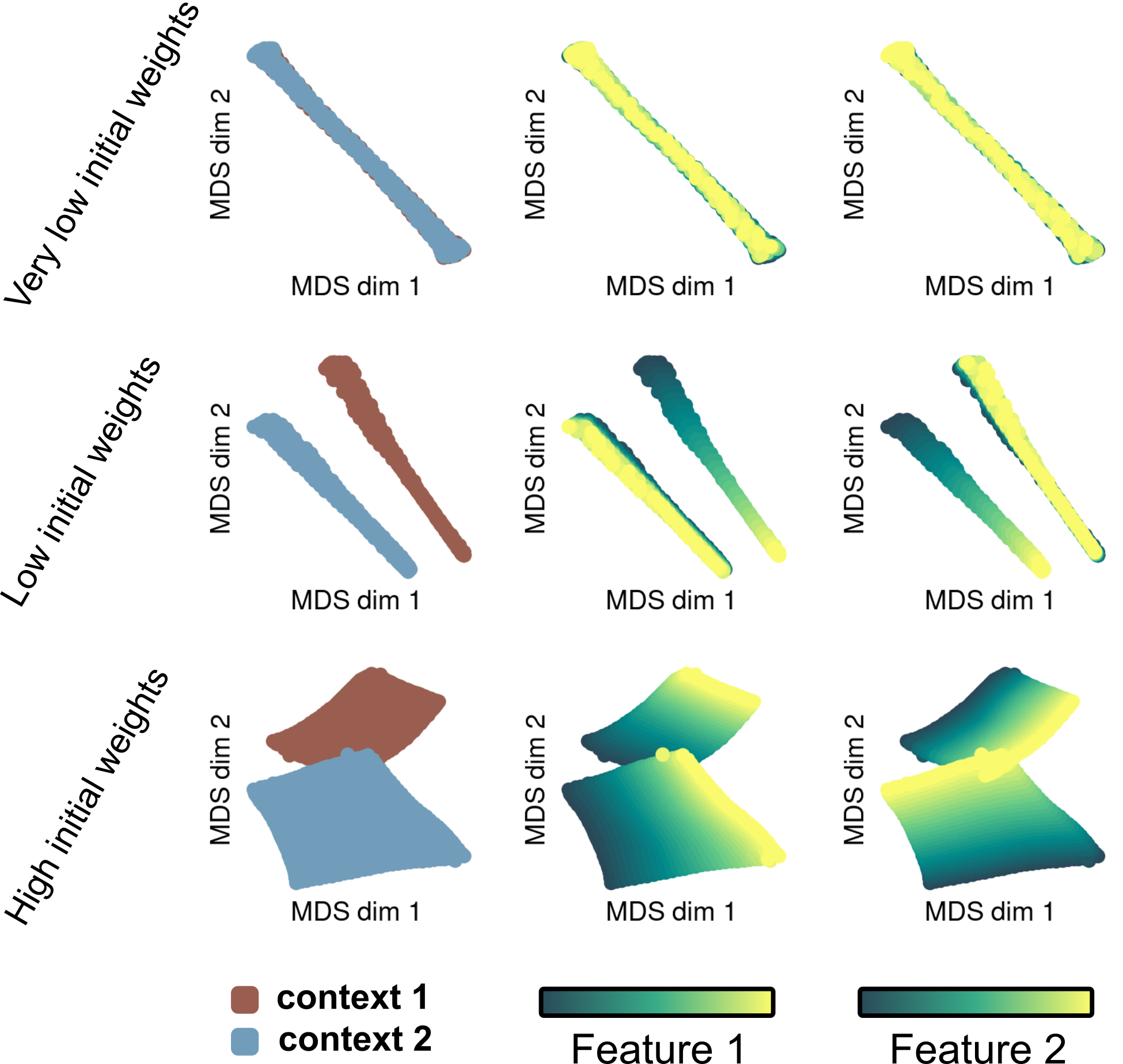}}
\caption{Representational structure in the orthogonal feature task, with Adam used as an optimizer. A multidimensional scaling projection of the representational structure of a 5-layer, tanh, fully connected network with 100 hidden units per layer, trained on the orthogonal feature dataset. At small initial weights the task-irrelevant direction collapses in each context, but remains at large initial weights.}
\label{fig:adam_feature_collapse}
\end{center}
\vskip -0.2in
\end{figure}

\begin{figure}[ht]
\vskip 0.2in
\begin{center}
\centerline{\includegraphics[width=0.4\columnwidth]{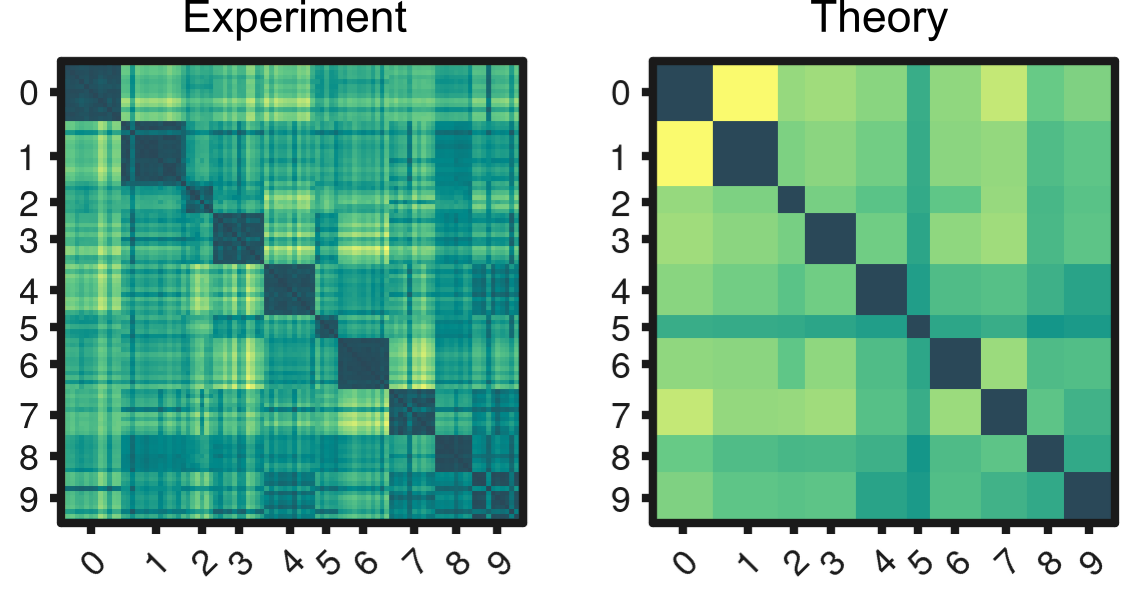}}
\caption{Representational structure on the MNIST dataset at small initial weights, with Adam used as an optimizer. Pairwise distances of the first 100 datapoints in MNIST after averaging over 50 trials of a 4-layer, leaky-ReLU, fully connected network trained on the full dataset (\textbf{left}) compared to the theory after exponential weighing (\textbf{right}).}
\label{fig:adam_mnist_dist}
\end{center}
\vskip -0.2in
\end{figure}

However, the intuitions discussed in \cref{sec:intuition} do not explicitly depend on some details of the optimization scheme, such as changes to the learning rate during training. Therefore, we may still expect a similar effect on the final representational structure. As can be seen in \cref{fig:adam_XOR}, \cref{fig:adam_feature_collapse} and \cref{fig:adam_mnist_dist} the representational predictions still hold up well.

\end{document}